\documentclass[preprint,12pt]{elsarticle}

\date{September 30, 2024}

\usepackage{color}
\usepackage{graphicx}
\usepackage{algorithm}
\usepackage{algpseudocode}
\usepackage{amsmath, amsthm, amssymb}
\usepackage{easyeqn}
\usepackage{subcaption}
\usepackage{url}

\usepackage{amssymb}
\usepackage{amsmath}

\def\beq{\begin{eqnarray}}
\def\eeq{\end{eqnarray}}
\def\be{\begin{EQA}[c]}
\def\ee{\end{EQA}}

\def\bm{\begin{math}}
\def\me{\end{math}}

\def\grad{\nabla}

\def\bel{\begin{EQA}[c] \label}

\newcommand \bei {\begin{itemize}}
\newcommand \eei {\end{itemize}}

\def\beel{\begin{eqnarray} \label}

\DeclareMathOperator*{\E}{\mathbb{E}}

\journal{CNSNS}

\begin{document}

\begin{frontmatter}



\title{Umbrella Reinforcement Learning -- computationally efficient tool for hard non-linear problems}


\author[sk]{Egor E. Nuzhin} 
\author[uol]{Nikolai V. Brilliantov\corref{cor1}}

\affiliation[sk]{organization={Applied AI, Skolkovo Institute of Science and Technology},
            addressline={Bolshoy Boulevard, 30, p.1}, 
            city={Moscow},
            postcode={121205}, 
            country={Russia}}
\affiliation[uol]{organization={Department of Mathematics, University of Leicester},
            addressline={University Road}, 
            city={Leicester},
            postcode={LE1 7RH}, 
            country={United Kingdom}}

\ead{nikolaigran@gmail.com}

\begin{abstract}
 We report a novel, computationally efficient  approach for solving hard nonlinear problems of  reinforcement learning (RL). Here we combine umbrella sampling, from computational physics/chemistry, with optimal control methods. The approach is  realized on the basis of  neural networks, with the use of policy gradient. It outperforms, by computational efficiency and implementation  universality, all available state-of-the-art algorithms, in application to hard RL problems with sparse reward, state traps and lack of terminal states. The proposed approach uses an ensemble of simultaneously acting agents, with a modified reward which includes the ensemble entropy, yielding an optimal exploration-exploitation balance. 
\end{abstract}


\end{frontmatter}




\section{Introduction}
Reinforcement learning (RL) is a rapidly developing field of Artificial Intelligence; it allows solving very different problems, ranging from purely engineering, to rather fundamental ones.  Numerous efficient RL algorithms have been proposed recently. Roughly, the  RL algorithms may be classified as model-free and model-based \citep{dong2020deep}, although other classifications are used \citep{wang2019benchmarking}.  Among the most popular  model-free algorithms are actor-critic class algorithms A2C \citep{mnih2016asynchronous} and PPO \citep{schulman2017proximal}. They demonstrated a great empirical success compared with other RL algorithms \citep{wu2020finite}. A number of model-free algorithms are based on  Q-learning \citep{watkins1992q} with various modifications, such as e.g. DQN \citep{mnih2013playing}, SAC \citep{haarnoja2018soft}, and DDPG \citep{lillicrap2015continuous}.  The algorithms  like iLQR \citep{li2004iterative}, MBMF \citep{nagabandi2018neural},  iLQR \citep{li2004iterative} and  MBMF \citep{nagabandi2018neural} are typical examples of model-based algorithms, see \citep{wang2019benchmarking} for a review. Here we refer to another class of algorithms (or algorithms' modifications), which were devised to tackle hard exploration or sparse reward  problems, see e.g. RND \citep{burda2018exploration}, ICM \citep{pathak2017curiosity}, Go-Explore \citep{ecoffet2021first} and  TR \citep{li2021solving} algorithms.

At the heart of the RL approach is the generation of state-action trajectories (episodes) associated with a particular agent policy. Utilizing the information, gained by the agent in an episode, the policy (that is, an action in response to a state) is improved for the next episode,  and the next trajectory is generated. Such an improvement of the policy continues until the best (optimal) policy is achieved\footnote{A number of trajectories  may be generated simultaneously, with the policy update based on several successive episodes.}. In spite of universality of such an approach, which may be deployed for numerous information-processing systems, including robots, advisory dialogue systems (e.g. ChatGPT\footnote{\url{https://openai.com/blog/chatgpt/}}), self-driving cars, etc., for some important for applications problems, RL may lose its efficiency and even fail. These problems are associated with (i) a delayed or sparse reward, (ii) lack of terminal states and (iii) state traps; we call them "hard problems" (see the discussion below). All these difficulties stem from the consecutive simulations. 

To overcome limitations caused by such simulations, one can apply methods of parallel simulations, that is, "whole-states" methods, similar to the  Value Iteration (VI) and Policy Iteration (PI) methods 
of classical dynamic programming \citep{howard1960dynamic}. 
Still, time discretization of the environment and the application of a discrete mesh of states and actions entail extensive computations for each update step of the action policy. Indeed, such an update is to be done for all  available state transitions. This can be extremely demanding, in terms of memory and CPU time, especially for high-dimensional state space -- the so-called "curse of dimensionalily"\citep{friedman1997bias,bengio2005curse}. 

Another class of RL approaches developed to tackle hard problems refer to the Demonstration Learning, which includes techniques like Behavior Cloning (BC) \cite{bain1995framework}, Inverse Reinforcement Learning (IRL) \cite{arora2021survey}, Offline Reinforcement Learning (Offline RL) \cite{levine2020offline}, and new approaches like Diffusion Policy \cite{chi2024diffusionpolicyvisuomotorpolicy} and Reverse Forward Curriculum Learning (RFCL) \cite{tao2024reverseforwardcurriculumlearning}. Here an expert demonstration allows an agent to learn a satisfactory policy  relatively fast,  even in challenging environments with sparse and delayed rewards, high-dimensional action spaces, and state traps. Unfortunately, the ability to incorporate expert knowledge directly into the learning process is not always possible; moreover, such methods do not allow to obtain  policy with a maximal reward. 

Next,  one should mention Exploration Bonus methods -- the RL  strategy  aimed to encourage agents to explore less familiar parts of the state space.  By providing additional intrinsic rewards, this  methods enable agents to discover states and actions, missed due to sparse rewards. They include:  Curiosity-Driven Exploration \cite{pathak2017curiosity}, Random Network Distillation \cite{burda2018exploration}, Count-Based Exploration  \cite{bellemare2016} and Policy Entropy Maximization  \cite{mnih2016asynchronous} and also suffer from significant limitations. In particular, these methods  heavily rely on the intrinsic rewards, which may not always align with the tasks associated with external goals. Furthermore, there is no universal recipe of construction intrinsic rewards -- it has to be done ad hoc for each problem. 

Similar techniques is the Reward Shaping, which may be exemplified by Intermediary Rewards for Low-Level Goals \cite{dorigo1998}, Potential-Based Reward Shaping \cite{ng1999} and Learning-Based Reward Shaping\cite{jaderberg2017}. 
 Here the reward is modified in order to provide more informative feedback to the agent. It is particularly valuable in environments where rewards are sparse, delayed, or difficult to obtain.  The reward reshaping  guides an agent toward the desired behavior and helps to accelerate learning. Nevertheless, although this technique can be highly effective, it possesses notable limitations. That is, 
designing an effective shaping rewards  requires at least  approximate knowledge of an optimal policy. A lack of such a knowledge would result in a poorly designed shaping functions yielding,  not optimal policy and/or unreasonably long  learning time.

Exploration-first methods, such as unsupervised exploration \cite{Laskin2020,hazan2019provably} strategies and systematic exploration methods like Go-Explore \cite{ecoffet2021first}, have been designed to address hard problems, which also contained a requirement for complex, long-term planning. Still, even an  exhaustive exploration (without optimized exploration-exploitation balance) cannot  guarantee that the optimal policy, with a maximal total reward will be found. 

Finally, one can mention another class of RL methods which address hard problems -- Intermediate goal-based approaches, such as Hindsight Experience Replay \cite{andrychowicz2017} and Hierarchical Reinforcement Learning \cite{nachum2018}. These methods introduce sub-goals or intermediate objectives that guide the agent's learning process, making it easier to reach the final goal. The deficiency of these methods stems from their high computational complexity and the need of designing appropriate   intermediate goals; these may be specific for particular problem. Hence, this approach is not universal.

Here we propose a new universal approach for solving hard problems of RL, which overcomes limitations of the consecutive simulations. It is computationally efficient and free from the "curse of dimensionality". Instead of dealing with a single agent or a finite number of agents, as in  conventional RL, it deals with a \textit{continuous} ensemble of agents (see a more formal definition below). The agents are characterized by a set of (state) variables, varying in some intervals. Such an ensemble is described by a distribution function for these variables, effectively corresponding to an infinite number of agents.\footnote{Note, that application of ensembles provides an additional flexibility of an approach, therefore  ensembles are effectively used in various settings, e.g. for the uncertainty estimation for Bayesian models \citep{lakshminarayanan2017simple}, transition probabilities evaluation \citep{dimakopoulou2018coordinated} , etc.} Conceptually, our approach is similar to 
Umbrella Sampling of classical Monte Carlo, utilized in Computational Physics and Chemistry. Hence  the term "Umbrella RL" is used.  

The rest of the article is organized as follows. In Sec. II we  define  hard problems for RL and discuss systems, where hard problems arise. In Sec. III we provide a detailed description of the algorithm of Umbrella RL and its practical implementation. In Secs. IV and V we demonstrate the application of the new algorithm to representative hard problems -- Multi-value Mountain  Car and StandUp problems. We also compare the performance of Umbrella RL with other, relevant for hard problems, RL algorithms and demonstrate a significant supremacy of the new approach. The final Sec. VI summarizes our findings.

\section{Definition of hard problems }
\begin{figure}[t]
     \centering
     \begin{subfigure}[b]{0.23\textwidth}
         \centering
         \includegraphics[width=\textwidth]{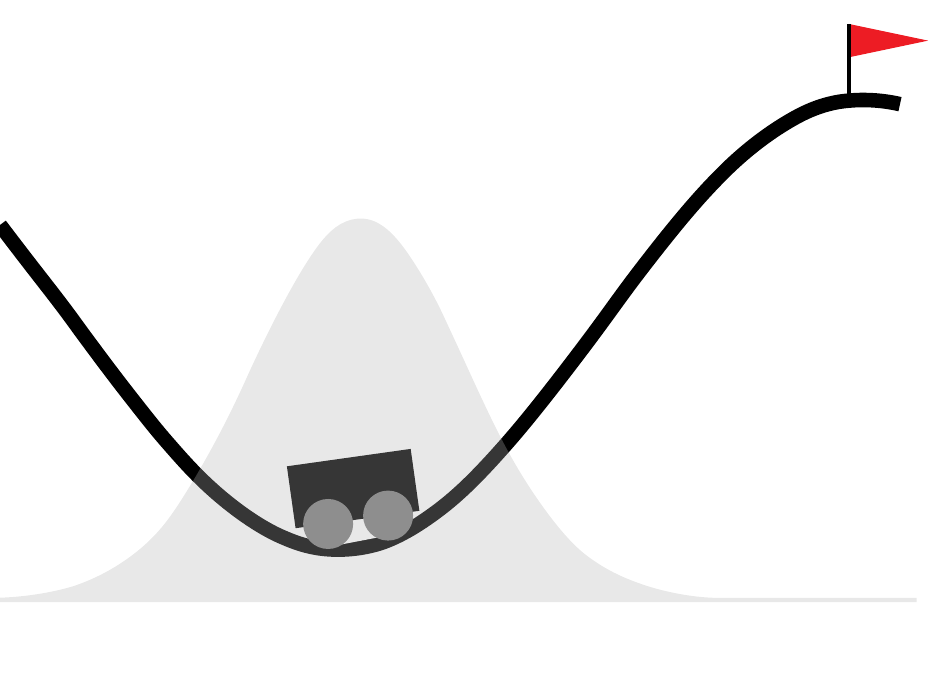}
         \caption{Mountain Car}
     \end{subfigure}
     \hfill
     \begin{subfigure}[b]{0.23\textwidth}
         \centering
         \includegraphics[width=\textwidth]{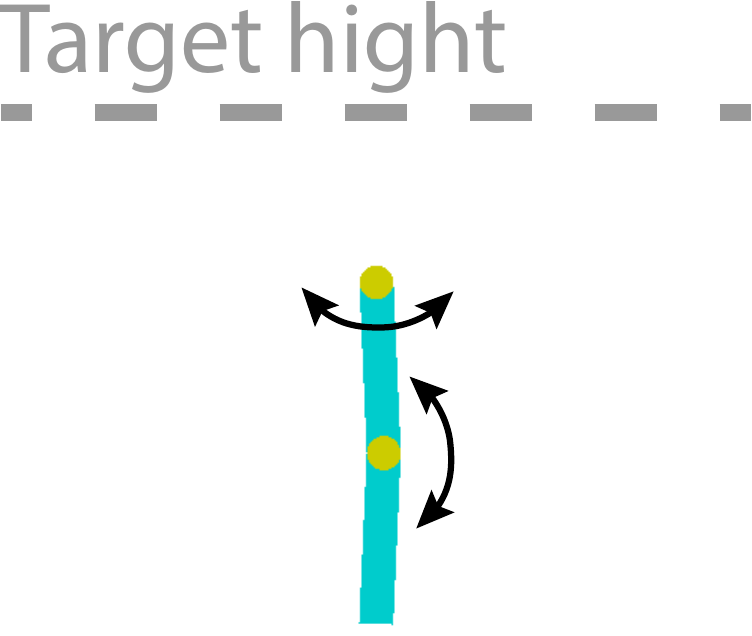}
         \caption{Acrobat}
     \end{subfigure}
     \hfill
     \begin{subfigure}[b]{0.23\textwidth}
         \centering
         \includegraphics[width=\textwidth]{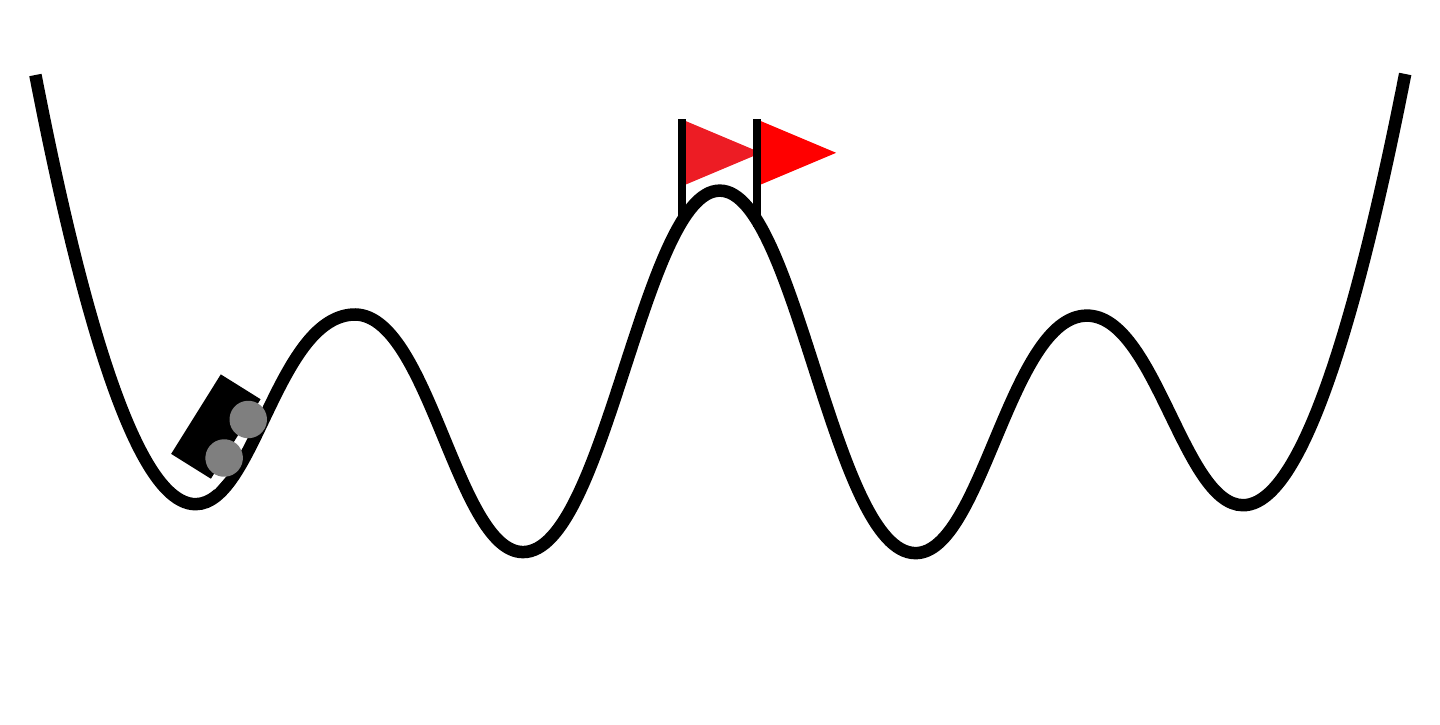}
         \caption{Multi-Valley Mountain Car}
     \end{subfigure}
        \hfill
     \begin{subfigure}[b]{0.23\textwidth}
         \centering
         \includegraphics[width=\textwidth]{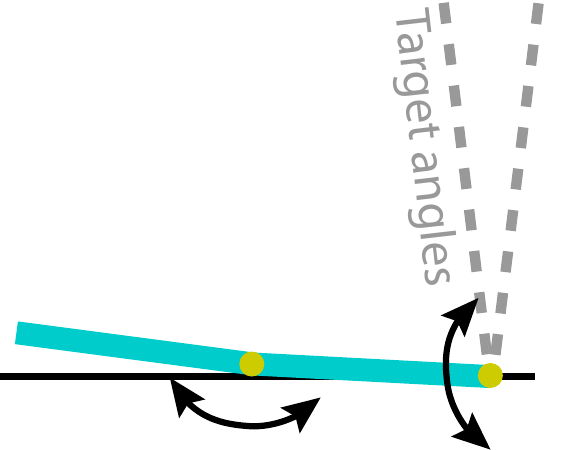}
         \caption{StandUp}
     \end{subfigure}
        \caption{Illustration of typical hard RL problems. Mountain Car (a) and Acrobot (b) problems are well-known for their complexity. Multi-Valley Mountain Car (c) and StandUp problem (d)  possess all main features of hard problems. The bell-shaped shadowed region in panel (a) illustrates the distribution of  ensemble of agents for Umbrella RL. }
        \label{fig:hard_prob}
\end{figure}
There are three main features of hard problems, where RL fails or dramatically loses its efficiency: 
\begin{itemize}
    \item A sparse, or significantly delayed reward; e.g. the reward may be gained only at the final state -- the goal.
    \item A presence of state traps, where an agent gets stuck, which are difficult to avoid. 
    \item A lack of a certain terminal state, e.g. the final goal may correspond to multiple states or a  state region.
\end{itemize}
The most prominent examples of hard problems are depicted in Fig. \ref{fig:hard_prob}. The Mountain Car problem \citep{moore1990efficient}, Fig. \ref{fig:hard_prob}a,  illustrates the first difficulty. Namely, the agent here is a car, randomly placed  in a valley. It may accelerate in either direction, which corresponds to possible actions, but cannot overcome the downhill force due to the gravity. The goal is to reach the top of the hill. This may not be generally  achieved by simple motion in one direction, but only by swinging in the well with an increasing amplitude, applying an appropriate acceleration in appropriate time instants. The top of the hill is the final position, and only there a reward is given. In other words,  any intermediate reward is lacking. This complicates an application of  RL which guides, by rewarding, towards a better (and eventually to the best) policy.  A simple random strategy --  a random acceleration here,  generally fails. 

Similarly, in the Multi-Valley Mountain Car problem,  the agent  is rewarded only reaching a zone between flags, see Fig. \ref{fig:hard_prob}c. Here the agent needs not only reach  the zone with the reward, but also balance within  the zone, without any terminal state.  Another hard RL problem is the Acrobot problem \citep{sutton1995generalization}, Fig. \ref{fig:hard_prob}b, where an "acrobat" needs to swing appropriately, to lift up to the target height, where the reward is given.  The same occurs in the StandUp problem, illustrated in Fig. \ref{fig:hard_prob}d -- the reward may be received at the very end of a successful action, that is,  in the target region. 

A safe evacuation from a  hazardous place, when a leader (e.g. a specialized robot) needs to guide a crowd  towards an exit, e.g.  \citep{bahamid2020intelligent}, belongs to the same class of problems. Saving lives is the main goal there; hence, a reward is given only  when all humans are evacuated. Otherwise the reward is  zero. Again,  there is no way to assess an efficiency of a strategy until the final state is reached. The list of examples may be continued. 

The second difficulty, of getting stuck in local state traps, is generic. Generally, it is associated with properties of the state-action space. For instance, when an agent gets stuck in a minimum of potential energy (like in Mountain Car), or when optimization of a policy drives it toward a local maximum of the expected return. 

State traps cause  a failure in walking of a humanoid robot -- here "get up to go" requires much more complex dynamics than simple walking straight \citep{kuindersma2016optimization}.

Finally, the third difficulty,  is related to the lack of a terminal state. It is also generic and arises in  numerous applications. For instance, when  learning certain movements, e.g. learning of an agent to walk  \citep{haarnoja2018learning}, or learning a defensive behavior by a swarm \citep{nuzhin2021animals}, a particular terminal state may be hard to define. Indeed, how is it possible to judge, whether an agent has already achieved  an efficient enough strategy? Such a conclusion seems to be very subjective. In other words, it is hard to define a terminal state in advance, that is, a clear criterion to cease simulations is lacking.  

Noteworthy, the above difficulties do not simply sum up, but enhance each other, making hard problems with a few difficulties even more challenging.  Fortunately, the novel approach, reported here, resolves  all the above problems. 

\section{The main ideas of Umbrella RL
} 

All the above hard problems difficulties   stem  from the consecutive nature of simulations, associated with  a generation of state-action trajectories for a singe agent, or a limited number of agents. What happens if one deals with a \textit{continuous} ensemble of agents, instead of discrete agents? By a continuous we mean an ensemble where the agents are characterized by a set of state variables of probabilistic nature, with the magnitudes varying  in certain intervals. It is described by a distribution function $p(s,t)$, where   $s$ is a multidimensional variable, specifying agents' states and $t$ is time. Correspondingly, $p(s,t)ds$ -- quantifies the number of agents with the state variable $s$ belonging to the small interval $(s,s+ds)$. 

The intention to use an ensemble is motivated by a somewhat  similar problem in computational physics and chemistry, where Monte Carlo (MC) method is used to compute thermodynamic properties.  In this case a direct application of Metropolis-Hastings algorithm may result in a stuck of a Markovian chain, associated with generated molecular configurations, in a local potential minimum. This prevents exploration of the whole  configurational space. The Umbrella Sampling MC provides an elegant solution to this problem: The Markovian chain is associated now, not with a single  thermodynamic state, but with a whole ensemble of different thermodynamic states, covering a certain region of thermodynamic phase space, as an umbrella  \citep{Umbr,FrenkelBook}. Hence, even  if some states  (cold and dense) act as a trap for a Markovian configurational chain, the presence of other  states of the ensemble (hot and rarefied) drive the Markovian chain  out of the trap, preventing  getting stuck. This allows a successful exploration of systems with complicated potentials and multi-phase systems, see e.g. \citep{Valleauinbook,Valleau2005,BV1998,BV1998a}.  

Here we adopt the main idea of Umbrella Sampling to RL, that is, we propose an approach where  an ensemble of agents explores simultaneously a whole range of states and actions. Hence the "umbrella" refers to the state-action space and the update of the policy is performed with the data accumulated from multiple realizations of the environment. For the same reasons as for Umbrella MC in Computation Science, such an algorithm will avoid the state traps. Moreover, Umbrella RL, additionally, resolves two other  difficulties of hard problems; this is demonstrated in the rest of the article. 

To this end, we need to design a method, where an ensemble of agents evolves in time, along with the evolution (towards the optimal) of the action policy. It should evolve in a way enabling  the  most efficient exploration of the environment. This would minimize time, needed to reach the desired states with a reward and help to avoid trapping.  Yet, the evolution of the ensemble should dramatically change when such states are reached. In terms of the exploration-exploitation trade off, we aim to construct an ensemble, that tends to be the most extended one, as long as states with a reward are lacking. Simultaneously, it should be the most condensed, once such states are reached -- condensed around the states with a reward. As we show below, this may be achieved, combining the conventional reward and entropy of the ensemble. Note, that the entropy of ensemble is joint state-action entropy and differs from  the  policy entropy \citep{ahmed2019understanding} generally used for regularization in RL.

\subsection{Policy gradient for Umbrella RL}
The policy gradient method has demonstrated its high efficiency and is incorporated in many RL approaches. Here we generalise this method, to deal  with a continuous  ensemble of agents. It is also convenient to consider continuous time. We apply the conventional assumption, that the system follows Markov decision process \citep{guo2009continuous,agarwal2019reinforcement}, so that agents make decisions (about actions) in accordance with a  random policy \(\pi\). The actions $a$ depends on the state only, that is, 

\begin{equation}
\label{1}
    a\sim \pi(a|s). 
\end{equation}
where \(s\) is the set variables characterizing an agent state. 

In other words, the action \(a\) is defined by the action policy \(\pi = \pi(a|s)\). The state of some  particular agent from the ensemble alters in time, subject to the evolution equation, 
\begin{equation}
\label{2}
    \dot{s} = v(s,a),
\end{equation}
where \(v\) is a given (multidimensional) function, quantifying the rate of change of a state with the (multidimensional) variable $s$. The above equations \eqref{1}-\eqref{2} govern the behavior of agents of the ensemble with the distribution function $p(s,t)$ and hence determine the evolution of  $p(s,t)$, with the  initial distribution
$p_0(s)=p(s,0)$.

The aim of RL, is to obtain an optimal policy \(\pi^*\), that is, the policy maximizing the total reward of the agent, which always acts in accordance with the  policy \(\pi\). For Markovian process it is determined  by $r(s,a)$ -- the reward the agent receives for an action $a$, made in a state $s$. The overall quality of a policy is quantified by the expected return,  $J(\pi)$  -- the  weighted sum of all rewards received on the  state-action trajectory. For Umbrella RL and continuous time it reads, 
\begin{equation}
\label{3}
   J(\pi) =  \int_0^\infty \gamma^t \left< \E_{a_t \sim \pi} \left[r(s_t,a_t) \right] \right>_{p_t}dt, 
\end{equation}
where $p_t$ abbreviates the distribution 
$p(s,t)$, depending on the policy $\pi$ and 
$\left<  \ldots \right>_{p_t} $ denotes the averaging over the distribution $p(s,t)$. Note that without this averaging  Eq. \eqref{3} is the continuous-time analogue of the common discrete-time {expected return}   \citep{sutton2018reinforcement}. 
The discount constant, \(\gamma \in [0,1)\), reflects the fact, that the immediate reward is more valuable, than that, delayed in time. 

Since the evolution of each agent from the ensemble $p(s,t)$  obeys Eqs. \eqref{1}-\eqref{2},  it may be shown that $p_t=p(s,t)$ obeys the equation,  
\begin{equation}
\label{4}
    \dot{p_t} + \nabla_s \left(  p_t \langle  v \rangle_{\pi}  \right)  =0, 
\end{equation}
which is, in fact, the continuity equation for the probability density and $\left<  \ldots \right>_{\pi } $ denotes  the averaging over the policy $\pi $ [see the Appendix,  Sec. \ref{sec:ap_st} for  detail].  Substituting the solution of Eq. \eqref{4}, $p(s,t)$,  into Eq. \eqref{3} one can find $J(\pi)$ for the given policy $\pi$. 

The optimal policy in the conventional RL is  found, by maximizing $J(\pi)$ with respect to the policy $\pi$. Obviously, such a policy optimization, for  the expected return, Eq. \eqref{3}, would hardly work for hard problems, where a reward is given only in a  small fraction of final states. Therefore,  the search for an optimal policy should be based on a generalized expected return, that contains  a term, stimulating  exploration of states. This term should dominate in the lack of states with reward, but be negligible, when such states appear. Based on this, the following generalized expected return is proposed:

\begin{equation}
\label{5}
    {\cal J}^{\rm URL}(\pi,p_t)=J(\pi) + \alpha \,  H[p_t \cdot \pi],
\end{equation}
where $J(\pi)$ is given by \eqref{3} and \(H[p_t \cdot  \pi]\) is the  joint entropy of the ensemble state distribution and action policy, 
\begin{equation}
\label{5a}
H[p_t\cdot \pi ] =  -\int_0^{\infty} \gamma^t\left< \E_{a_t \sim \pi} \left[\log (p_t \cdot \pi)\right] \right>_{p_t} dt 
\end{equation}
and \(\alpha\) is the entropy weight.
 
Noteworthy, $H[p_t\cdot \pi ] $  is similar to the conventional entropy based on the trajectories [see e.g. \citep{TrEnt}],  but with the time discount factor $\gamma$ and  probability \(p_t\) under the logarithm. In other words, this entropy is the generalization (with the discounted reward and for  continuous time)  of the state-action entropy \citep{neu2017unified} used in entropy-regularized MDP.
For simplicity we will call the modified entropy \eqref{5a} as entropy. Correspondingly, the optimal policy is defined  by the following optimality condition;

\begin{equation}
\label{7}
\pi^* 
= \arg \max_\pi \left\{J(\pi) + \alpha  H[p_t \cdot \pi]\right\}.
\end{equation}
As it follows from the above equation, in the absence of  states with reward, $J(\pi)= 0$, 
and the optimal policy maximizes the entropy only.

The maximization of the entropy, in its turn, implies the widening of the distribution $p(s,t)$. As the result, more and more states get covered during the optimization procedure. At some optimization stage  the states with a reward  become included in $p(s,t)$ and $J(\pi)$ becomes non-zero. If the coefficient $\alpha$ is small, $J(\pi)$ in ${\cal J}^{\rm URL} (\pi,p)$ starts significantly  dominate   and the optimization procedure follows that of  conventional RL. 

In practice, it is more convenient to work with a simplified entropy, based not on the bunch of trajectories, as in Eq. \eqref{5a}, but rather on a single distribution, $\overline{p}(s)$, found by averaging  the distributions $p(s,t)$: 

\begin{equation}
\label{pav}
    \overline{p} (s) = \int_0^\infty \gamma^t p(s,t) dt.  
\end{equation}
Here we omit the normalization factor $|\log \gamma|$ as it will not affect the subsequent analysis. With the average distribution $\overline{p}(s)$, the simplified entropy reads, 
\begin{equation}
\label{Hdef1}
    H[\overline{p}\cdot \pi ] = -\left< \E_{a_t \sim \pi} \left[\log \left(\overline{p}\cdot \pi  \right) \right] \right>_{\overline{p}},
\end{equation}
which significantly decreases  the computational complexity, especially in terms of required memory.

The search of the optimal policy $\pi^*$ is usually performed within a convenient class of functions with an appropriate  parametrization. 

This reduces the problem  to that, of finding the optimal state-dependent,  multi-component parameter $\theta^*$, associated with $\pi^*$, that is, $\pi^*= \pi_{\theta^*}$.  The value of  $\theta^*$ may be found from the policy gradient  $\grad_\theta J(\pi_\theta)$, expressed in terms of the policy and advantage function, \(A_{\rm u}(s,a)\), modified for Umbrella RL:

\begin{equation}
\label{8}
     \grad_\theta {\cal J}^{\rm URL}(\pi_\theta) =  \left<  \E_{a \sim \pi } \big[\grad_\theta \log  \pi_\theta(a|s) \cdot A_{\rm u}(s,a)\big]  \right>_{\overline{p}}.
\end{equation}
where \(A_{\rm u}(s,a)\) is defined below\footnote{Here we again omit unimportant factor $|\log \gamma|^{-1}$.}, see Appendix  (Sec. \ref{sec:ap_dif_pg}) for the derivation detail,  which follows the standard approach, generalized for the continuous variables, e.g. \citep{sutton2018reinforcement}.

Note, that in contrast to conventional RL, the gradient of  the modified expected return depends on the averaged ensemble distribution $\overline{p}$.  Similarly, the modified advantage  also depends on this distribution. The derivation of $A_{\rm u}(s,a)$  is also straightforward and detailed in Appendix (Sec. {\ref{sec:ap_dif_pg});  here we provide the final result, 

\begin{equation}
\label{9}
A_{\rm u}(s,a) = r_u+v \cdot \nabla_s V_{\rm u}(s)   - |\log {\gamma} | \,  V_{\rm u}(s),  
\end{equation}
where we abbreviate $r_u=r-\tilde{\alpha} \log \big(\overline{p} \cdot \pi\big)$, which is the effective reward, with $\tilde{\alpha}= \alpha |\log \gamma|$ and  define the value function $ V_{\rm u}$, generalized for the case of Umbrella RL:
\begin{equation}
\label{10}
    V_{\rm u}(s)= \E_{a \sim \pi}  \left\{ \int_0^\infty \gamma^t  \left[ r_u(s_t,a_t) \right]_{s_0=s} dt \right\}.
\end{equation}

Using the value of the policy gradient $\grad_\theta {\cal J}^{\rm URL}(\pi_\theta)$ for some (multi-component) parameter $\theta_0$, one can update it according to the equation
$$
{\bf \theta}_1 =\theta_0 + \beta^{\pi}  \,  \grad_\theta {\cal J}^{\rm URL}(\pi_{\theta_0}),
$$
where $\beta^{\pi} > 0 $ is the  so called learning rate used for the policy upgrade. $\theta_1$ gives rise to the new, improved policy $\pi_{\theta_1}$, used to compute the next policy gradient for the further improvement  of the policy. Generally, 
\begin{equation}
\label{11}
{\bf \theta}_{k+1} =\theta_k + \beta^{\pi} \,   \grad_\theta {\cal J}^{\rm URL}(\pi_{\theta_k}). 
\end{equation}
Such a procedure continues until the optimal policy with  $\grad_\theta {\cal J}^{\rm URL}(\pi_{\theta_k}) =0$ is achieved, where $\theta_{k+1}=\theta_k=\theta_*$. In practice, the computations proceed  until $|\theta_{k+1}-\theta_k| < \varepsilon$, where $\varepsilon$ is small.

\subsection{Effective computation of the policy gradient}

To compute ${\cal J}^{\rm URL}(\pi_{\theta_k})$, given by Eq. \eqref{8}, one needs to find,  for each step $k$, the average distribution $\overline{p}(s)$, from Eq. \eqref{pav} and the  advantage function,  $ A_{\rm u}(s,a)$, defined by Eqs. \eqref{9}-\eqref{10}. The straightforward computation of these quantities is extremely inefficient and can hardly have practical application. Here we present computationally effective approaches to find $\overline{p}(s)$ and $A_u(s,a)$. 

It may be shown [see Appendix (Sec. \ref{sec:ap_value})], that the advantage function obeys the equation, 
\begin{equation}
\label{12}
    \E_{a \sim \pi} A_{\rm u}(s,a) = \overline{v}\cdot \nabla_s V_{\rm u}(s)  + \overline{r}_u - |\log {\gamma} | \,  V_{\rm u}(s)  =0, 
\end{equation}
where we introduce averages, $\overline{r}_u \equiv  \E_{a \sim \pi} \left[r_u (s,a) \right]$ and $\overline{v} \equiv  \E_{a \sim \pi} \left[v (s,a) \right] $.  Solving Eq. \eqref{12} for $V_{\rm u}(s)$ and substituting it into Eq. \eqref{9} we find $A_{\rm u}(s,a)$. 
Similarly, instead of using Eqs. \eqref{4} and \eqref{pav}, the average ensemble density $\overline{p}(s)$ may be found as the solution of the following equation [see Appendix (Sec. \ref{sec:ap_ss_for_p})
for detail]: 
\begin{equation}
\label{14}
       \nabla_s\cdot \overline{p}(s) \overline{v} -\log \gamma \, \left(\overline{p}(s) -p_0(s) \right)  = G\left(\overline{p}(s),s \right) =0, 
\end{equation} 
where $p_0(s)=p(s,0)$ and we introduce the growth rate function,
\begin{equation}
\label{14a}
       G\left(\overline{p}(s),s \right) = \nabla_s\cdot \overline{p}(s) \overline{v} -\log \gamma \, \left(\overline{p}(s) -p_0(s) \right),  
\end{equation} 
which steady-state vanishes, $G\left(\overline{p}(s),s \right) =0$. The solution of the PDEs  \eqref{12} and \eqref{14} may be performed using neural networks (NN). This yields a dramatic increase of the computational efficiency, as compared to the direct calculation of $V_u(s)$ and  $\overline{p}(s)$ by Eqs. \eqref{10} and \eqref{pav}. Moreover, Eqs. \eqref{12} and \eqref{14} may be solved  again,  using the policy gradient method, see Appendix (Sec. \ref{sec:ap_ass}).

\begin{algorithm}
\caption{Umbrella RL}
\label{alg:ur}
\textbf{Input}: Initial policy, value function and  average agents distribution parameters: \(\theta_0\),  \(\phi_0\) and \(\eta_0\); discount factor \(\gamma\). Here $\theta_0$, $\phi_0$ and $\eta_0$ are taken as random numbers from the interval $[-1/\sqrt{f},1/\sqrt{f}]$, where \(f\) is the number of input features  for a layer (default PyTorch initialization) \citep{lecun2002efficient}\\
\begin{algorithmic}[1]
\For{k = 0,1,2...}
\State  Sample a set of states \(\{s_1, s_2,...,s_{D_k}\}\) from an arbitrary probability distribution \(\rho(s)\). Here \(D_k\) is the number of samples for the step number  \(k\). 
\State Sample actions \(\{a_1, a_2,...,a_{D_k}\}\) by the policy \(\pi\) corresponding to each state \(s \in \{s_1, s_2,...,s_{D_k}\}\).
\State Evaluate: initialization probability  \(p_0\); reward \(r\), state  rate  \(v\); rate divergence \(\grad_s \cdot v\) for each state-action pair.
\State Using the value function \(V_{\phi_k}\) and its state gradient \(\grad_s V_{\phi_k}\), compute the advantage \(A_i \equiv A_u(s_i)\), defined by Eq. \eqref{9}, for each $i \in [1,D_k]$.
\State Using the average probability distribution \(\overline{p}_\eta (s) \big|_{\eta_k}\) and its state gradient \(\grad_s \overline{p}_\eta (s) \big|_{\eta_k}\) compute the growth rate  \(G_i \equiv G(\overline{p}(s_i),s_i)\), defined by Eq. \eqref{14a} for each $i \in [1,D_k]$.
\State Evaluate the estimates of the stochastic gradients:
\begin{equation}
\begin{split}
\label{AGLeq}
    \hat{g}_k^\pi &= \frac{1}{D_k}  \sum_{i = 1}^{D_k}  \nabla_\theta \log \pi_\theta(a_i|s_i)\big|_{\theta_k}  A_i ,\\
     \hat{g}_k^V &= \frac{1}{D_k}   \sum_{i = 1}^{D_k} \nabla_\phi V_\phi (s_i) \big|_{\phi_k}  A_i ,\\
      \hat{g}_k^p &= \frac{1}{D_k}   \sum_{i = 1}^{D_k} \nabla_\eta \log \overline{p}_\eta (s_i) \big|_{\eta_k}  G_i.
\end{split}
\end{equation}

\State Update the NNs, using the gradient ascents:
\begin{equation}
\begin{split}
    \theta_{k+1} &= \theta_k + \beta^\pi \hat{g}^\pi_k,\\
    \phi_{k+1} &= \phi_k + \beta^V \hat{g}^V_k,\\
    \eta_{k+1} &= \eta_k + \beta^p \hat{g}^p_k,
\end{split}   
\end{equation}
or using any other gradient algorithm (e.g. Adam).
\EndFor
\end{algorithmic}
\end{algorithm}
\subsection{Algorithmic implementation of Umbrella RL}
Neural networks (NNs) allow a very efficient computational implementation of Umbrella RL. This may be done with three NNs --  for the desired policy $\pi(a|s)$, for the expected reward $V(s)$ and for average agent distribution $\overline{p}(s)$. Let these three NNs be parametrized, respectively, by the multi-component parameters $\theta$, $\phi$ and $\eta$. Our goal is to find the extremal values of these parameters $\theta^*$, $\phi^*$ and $\eta^*$. Here $\theta^*$ corresponds to the optimal policy, $\pi^*$, which is a steady solution ($\theta_{k+1}=\theta_k=\theta^*$) of the difference equation \eqref{11}. $\phi^*$ and $\eta^*$ correspond to the values, which guarantee that the parametrized functions $V(s)$ and  $\overline{p}$ satisfy  Eqs. \eqref{12} and \eqref{14}.  They are also the steady solutions of the difference equations, \eqref{phik}--\eqref{etak} given below, which may be obtained, applying the policy gradient method [see Appendix (Sec. \ref{sec:ap_ass})]. 
Hence,  all three equations for $\theta$, $\phi$ and $\eta$ may be written in a uniform way,  with the  learning rates, $\beta^V$, $\beta^p$ and $\beta^{\pi}$,

\begin{eqnarray}
\label{phik}
{\bf \phi}_{k+1}=  \phi_k + \beta^{V} \,  \left< \E_{a \sim \pi} \left[\grad_{\phi}{V^\pi (s)} \cdot A(s,a)\right] \right>_{\overline{p}}, \,\,\,\,\,\,\,\,\,\,\,\,\\ 
\label{etak}
{\bf \eta}_{k+1} =  \eta_k + \beta^p \left< \E_{a \sim \pi} \left[\grad_{\eta}{\log \overline{p}(s)} \cdot G(\overline{p}(s),s) \right]\right>_{\overline{p}},\,\,\,\, \\
\label{thetak}
{\bf \theta}_{k+1} = \theta_k + \beta^{\pi} \,  \left<  \E_{a \sim \pi } \big[\grad_\theta \log  \pi(a|s) \cdot A_{\rm u}(s,a)\big]  \right>_{\overline{p}},
\end{eqnarray}
which manifests the possibility of applying the same computation algorithm. Note that the last equation \eqref{thetak} corresponds to the previous  Eq. \eqref{11}. In practice,  one can make averaging $\left< \cdots \right>_{\rho}$ in  Eqs. \eqref{phik} - \eqref{thetak} over any distribution \(\rho(s)\), covering the available observation range, that is, the averaging $\left< \cdots \right>_{\overline{p}}$ may be substituted with $\left< \cdots \right>_{\rho}$. Using the uniform distribution for \(\rho(s)\) drastically simplifies computations.  

The above algorithm with three NNs may be straightforwardly implemented  with modern auto-differentiation packages such as PyTorch\footnote{\url{https://pytorch.org}} and TensorFlow\footnote{\url{https://www.tensorflow.org}} (see the implementation on GitHub\footnote{The implementation of the UR algorithm,  Multi-valleys MC and SatndUP  environments are available on GitHub: \url{https://github.com/enuzhin/ur}}).
In this implementation, the value function \(V_u(s)\) and its derivative \(\grad_s V_u(s)\) is approximated by the first network. It is then  used to compute the advantage,  \(A_u\),  needed for the value function and policy update. The second network approximates the average probability density $\overline{p}$ and its derivative.
 The last network calculates logit parameters of the agent policy $\pi$. Computation of the derivatives of the functions \(\grad_s V_u(s)\) and \(\grad_s \overline{p}(s)\) is needed as they are used in the update procedures. The Algorithm \ref{alg:ur} presents  the algorithmic implementation of the Umbrella RL.

 \section{Application of the Umbrella RL}
To illustrate a practical application of the general ideas of the Umbrella RL, we analyze a few hard RL problems. We have chosen such of them that combine all the difficulties of the hard problems and demonstrate that Umbrella RL can excellently solve them. 

\subsection{Multi-Valleys Mountain Car problem}
\label{sec:mvmc}

In the  Multi-Valley version of the  Mountain Car problem, a central (highest) hill separates a number of hills and valleys to the left and to the right from the central hill, see Fig.  \ref{fig:hard_prob}c. The agent receives the reward only in a narrow range of positions (between two flags) on the top of the central hill.  The initial location is  possible in either of the two outermost valleys. The multi-valley version is a harder problem than its basic, one-valley, counterpart. The most important complication here is that it is not enough just to climb the hill. Firstly, it should be the central hill, secondly,  once arriving at the top of the central hill, the agent should balance  there, keeping  speed within the safe range. Such a scenario of the multi-valleys problem lacks termination conditions, which is generic and may be hardly treated by standard RL algorithms. 

The problem is formulated in terms of two state variables specifying the agent (car) position $x$ and its velocity $\dot{x}$. The environment is determined by the current  height $y(x)$, 
\begin{equation}
\label{height}
    y(x) \!= \! \frac{1}{10} \! \left[\cos \left(2 \pi x\right)\! + \! 2 \cos \left(4 \pi x\right) \!- \!\log \left(1 - x^2\right)\right],
\end{equation}
which depends on the car position $x$. 

The interaction with the environment occurs through the actions of the agent, applying one of the two discrete actions,  \(a \in \{0,1\}\) - the acceleration to the left, $a=0$, or to the right, $a=1$. The equation of motion reads,

\begin{equation}
\label{yx}
    \ddot{x} = (2 \cdot a - 1) \cdot f - y'(x) \cdot g.
\end{equation}

Here \(f = 0.001\) determines the agent's force,  \(g = 0.0025\) is the gravitational acceleration and $y'(x) \cdot g$ is its component driving the car downhill. The agent is rewarded,  when it resides between the flags, \(x  \in [-0.05,0.05]\),  with the reward \(r = 1\); otherwise the reward is zero,  \(r = 0\).

Evolution of the state function $s=\{x, \dot{x} \}$ obeys the equation, 
\begin{equation}
\label{sdotv}
\dot{s} \!=\! \left( \dot{x}, \ddot{x} \right)^T \!= \!v(s,a) \!=\! \left(\dot{x},\, \, \, \left(2a - 1\right) f - y'(x) g \right)^{T}.
\end{equation}

Eqs. \eqref{height} and \eqref{sdotv} should be supplemented by the initial condition. We use the following one: $p_0(x, \dot{x})=q_0$ if $x \in [-0.77,-0.67] \cap [0.67,0.77]$ and $\dot{x} \in [-0.01,0.01]$, and $p_0(x, \dot{x})=0$ otherwise. Here $q_0=250$, follows from the normalization, \(\int p_0(x,\dot{x}) dx d\dot{x} = 1\). 

Eqs. \eqref{height} and \eqref{sdotv} with the initial conditions completely define the problem, so that the above Umbrella RL algorithm, Eqs. \eqref{phik} -- \eqref {thetak} may be applied. Noteworthy, we compute \(G_i\) in Eqs. \eqref{14a}, \eqref{AGLeq}, using  the special relation \( \nabla_s\cdot p  \overline{v} =  p \E_{a \sim \pi}\left[ v \cdot \grad_s{\ln \left(\pi \cdot p\right)}\right] \), which may be straightforwardly derived (see Appedix, Sec.  \ref{sec:ap_state_div} for derivation detail). 

\subsection{StandUp problem}
\label{sec:ap_standup}

The problem is designed to teach a roboarm lying on the plane to lift up and balance in the vertical position, see Fig. \ref{fig:hard_prob}d. The arm consists of two bars connected by an active joint. One end of the roboarm is free and the other one is fixed to the plane with another active joint. Each of the active joints may exhibit a torque in either direction (clockwise and counterclockwise). We consider the overdamped case of large friction torque, when inertial effects may be neglected. The goal of the roboarm is to apply torques on actuated joints to achieve a balance in the straiten-up vertical position. The maximal torque of the joints is small and does not allow to lift up the straightened configuration of the roboarm; this can be done only for the folded one. 

The actions are discrete and represent the torque applied between the arm joints. We define for the problem four possible actions \\ \(a \in \{[-m,-m],[-m,m],[m,-m],[m,m] \)\}, where \(m = 0.0375\) is a fixed constant torque. The first  quantity in the square brackets here is a torque \(m_1\) of  the joint fixed to the plane, while the second one,  \(m_2\), is a torque of the joint connecting the bars. The observation space consists of two rotational joint angles (see Fig. \ref{fig:hard_prob}d):

\begin{itemize}
    \item \(\phi_1\) -- is the angle between the plane and the bar linked to the fixed joint (the first bar).  Zero angle indicates the first bar directed exactly rightwards.
    \item  \(\phi_2\) -- is the angle between the first and the second bar with the free end.  Zero angle indicates a fully straighten arm.
\end{itemize}
The angles are measured in the counterclockwise direction. The reward is assigned when the roboarm is straightened up, perpendicular to the plane. Some deviations from the vertical direction and the straight configuration are allowed -- the agent is rewarded with $r=1$, when it resides in the goal states range \(\phi_1^g \in (\pi/2 -\Delta,\pi/2 + \Delta)\) and \(\phi_2^g \in (-\Delta,\Delta)\), with \(\Delta = \pi/24\), otherwise $r=0$.

The episode starts with random initialization of the agent  angles. Roboarm lies on one of the possible sides: left or right with equal probability. Being on the right, the initial angles are distributed as \(\phi_1^r  \in (0, \Delta)\) and \(\phi_2^r \in (0, \Delta)\), with the similar initial distribution being on the left.  Assuming the unit mass and bars length, the over-damped equations of motion read, 

\begin{eqnarray}
    \label{phi1} 
    \dot{\phi}_1 &=& m_1-m_2 - g \cos \phi_1\\ 
    \label{phi2} 
    \dot{\phi}_2 &=& m_2 - g \cos (\phi_1+\phi_2).
\end{eqnarray}
Here \(g = 0.025\) is the gravity acceleration and the agent action $a$  is the two torques, \( a \equiv \{m_1,m_2 \} \). Eqs. \eqref{phi1}-\eqref{phi2} are applicable for \(\phi_1 \in (0,\pi)\) and \(\phi_2 \in (-\pi,\pi)\), with the  constraints of impenetrable plane, \(\phi_2 \in (-2\phi_1 , 2 \pi - 2\phi_1)\).

Numerically, we apply angle clipping to satisfy the constraints. 
Note, that we compute \(G_i\) in Eqs. \eqref{14a}, \eqref{AGLeq}, using  the relation \\ \( \nabla_s\cdot p  \overline{v} =  p \E_{a \sim \pi}\left[\grad_s \cdot v + v \cdot \grad_s{ \ln \left(\pi \cdot p\right)}\right] \), see Appendix (Sec.  \ref{sec:ap_state_div}) for derivation detail.

\subsection{Numerical experiments. Implementation detail }
\label{sec:td}
We performed numerical experiments  using  the above Umbrella RL algorithm with the feed-forward neural networks. We utilized three NNs to obtain the average density  distribution \(\overline{p} (s)\), value-function \(V_u(s)\) and action-policy \(\pi(a|s)\). Each NN had three layers, comprised of 128 neurons. We used ELU and TanH activation functions combined with NN layers,  according to Table \ref{table:ur_arc}; the Adam algorithm was applied. The number of iterations was $1.2 \cdot 10^6$, the batch size -- $10^4$, the entropy coefficient $\tilde{\alpha} =10^{-2}$ and the discount factor $\gamma = 0.95$. A uniform sampling was chosen. 

Umbrella reinforce algorithms utilized different learning hyperparameters for Multi-Valley MC and StandUp problems. We specified for Multi-Valley MC learning rate of \(10^{-5}\) for all three NN of the Adam algorithm. The weight decay parameters were \(10^{-4}\) for the values function NN, \(5\cdot10^{-4}\) for distribution NN and \(5\cdot 10^{-6}\) for policy NN. Similarly, for StandUp problem, the learning rates were \(10^{-6}\) for value function NN, \(10^{-7}\) for the distribution NN and \(10^{-6}\) for policy NN. The respective weight decay parameters were \(10^{-5}\), \(5\cdot10^{-4}\) and \(5\cdot 10^{-5}\).

\begin{table*}[!ht]
\scriptsize
\centering
\begin{tabular}{||c | c||} 

 \hline
 \(V_u(s)\) & Repre-n \(\rightarrow\) Linear(\(n_r\),128)\(\rightarrow\) ELU \(\rightarrow\) Linear(128,128) \(\rightarrow\) ELU \(\rightarrow\) Linear(128,1)  \\[1ex]  
   \(p(s)\) &  Repre-n \(\rightarrow\) Linear(\(n_r\),128)\(\rightarrow\) ELU \(\rightarrow\) Linear(128,128) \(\rightarrow\) ELU \(\rightarrow\) Linear(128,1) \(\rightarrow\) Exp   \\[1ex]  
  \(\pi(a|s)\) & State \(\rightarrow\) Linear(2,128)\(\rightarrow\) TanH \(\rightarrow\) Linear(128,128) \(\rightarrow\) TanH \(\rightarrow\) Linear(128,2)  \\[1ex]
 \hline
\end{tabular}
\caption{Neural network architectures for Umbrella RL}
\label{table:ur_arc}
\end{table*}

In Multi-Valley Mountain Car computations we limited the computation domain of states, such that $x \in [-0,99,099]$
 and $\dot{x}=[-0.07,0.07]$ and apply reflective boundary 
 conditions, see Appendix (Sec. \ref{sec:ap_bc_mvmc}). For StandUp problem, the boundary conditions follow from the physical constraints, see Appendix (Sec. \ref{sec:ap_bc_standup}).

To assess the efficiency of our approach, in comparison with other methods,  we  perform experiments using state-of-the-art RL algorithms. Moreover, among them we have chosen  the algorithms, expected to be the most efficient for hard problems. Namely, we apply the widely used RL algorithm of proximal policy optimization (PPO); it was also claimed to be the most effective and universal \citep{schulman2017proximal}. We make experiments with value iteration (VI) algorithm -- the fundamental algorithm of RL.  We also use PPO algorithm with the   prominent exploration bonus, RND,  and iLQR -- the popular model-based RL algorithm. Note, that as it was shown in \citep{burda2018exploration}}, RL with  RND noticeably outperforms RL with  curiosity-driven exploration bonus \citep{pathak2017curiosity}.  

Additionally, we checked the efficiency of the incomplete Umbrella RL (called UR-NE), with  ${\cal J}^{URL}$  substituted by $J$, without the entropy term. 

The PPO algorithm was applied from Stable Baselines3\footnote{\url{https://stable-baselines3.readthedocs.io/}} package. We used default parameters and trained the network for the same number of iterations,  as for UR, equal to \(1.2 \cdot 10^6\); we made the environment reset every \(2 \cdot 10^3\) steps. 
The Value Iteration algorithm was implemented from scratch. It was applied on a discrete equidistant grid of 9000x9000 size. The training was performed until the convergence. The value function was initialized by  zeros.  The RND algorithm\footnote{\url{https://github.com/jcwleo/random-network-distillation-pytorch}} was applied for the problems where the observation was a sequence of images. It was used with default parameters \citep{burda2018exploration} for the \(10^7\) iteration in total. Due to the algorithm restrictions, with respect to discontinuities, we apply the iLQR algorithm\footnote{\url{https://github.com/Bharath2/iLQR}}, 
 using smoothing (with a Sigmoid) of the involved functions. A trajectory of 10 000 steps has been optimized, until iLQR convergence.  We evaluated the agent policy performance in several individual simulation with fixed time-discretization step.  The results were averaged and the variance estimated.

\section{Results for the hard problems}
\label{sec:res}
\subsection{Umbrella RL for Multi-Value Mountain Car problem}

\begin{figure}[t]
\centering
\includegraphics[width=0.875\columnwidth]{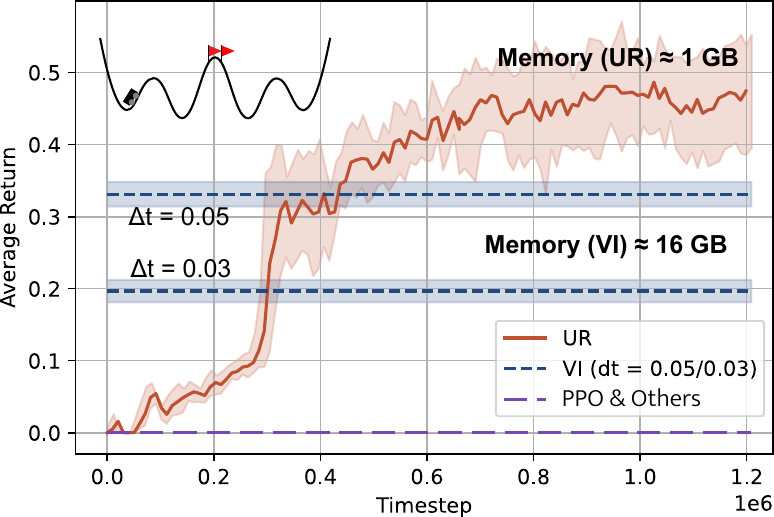}
\caption{Average expected return for Multi-Valley Mountain Car problem: For UR, UR-NE, PPO, RND, and iLQR the simulation time step is \(dt = 0.05  \) in each training run of 50 episodes (10 runs in total). For VI algorithm additional time steps of $0.03$ and $0.01$ are included.  The total simulation time is 100. The "Others" algorithms on the figure -- UR-NE, RND, iLQR  (dt = 0.05) and VI (dt = 0.01) obviously fail. }
\label{fig:act_per_mvmc}
\end{figure}

In Fig. \ref{fig:act_per_mvmc} the expected return  as the function of training time is presented.   It is clearly seen from the figure that the new Umbrella RL algorithm significantly outperforms all the alternatives:  While PPO, RND, iLQR, and  UR-NE completely fail (yielding almost zero cumulative reward), our algorithm exceeds VI  in terms of discounted cumulative reward.  It is also obvious that the entropy term in the  expected return  is crucially important -- the algorithm loses the efficiency if this term is lacking. One can also conclude that the traditional, trajectory-based algorithms, such as PPO,  fail for this hard problem. In contrast, all-states approaches, as VI, or ensemble-based approach, as the new Umbrella RL, demonstrate the ability to solve the hard problem. Still, the VI approach strongly relies on time and state discretization. For a complex environment, the amount of memory, needed to keep an acceptable accuracy drastically increases. This results in a substantial decrease in performance. As it may be seen from Fig.  \ref{fig:act_per_mvmc}, the efficiency of the  VI remains comparable (although smaller) with  the Umbrella RL for the time step, $dt=0.05$, it noticeably drops for $dt=0.03$ and becomes unacceptable for $t=0.01$ and smaller time steps. Here the time step of \(dt = 0.05\) suffices for the studied environment with  a rather small gravity, $g=0.0025$; the increasing $g$ would require the decreasing time step, making the VI inappropriate. 

The main advantage of the Umbrella RL, with respect to VI,  is independence of the former on the discretezation timestep. 
Although the number of iterations for VI to reach the convergence is smaller than that for UR, the required time per iteration and memory usage is significantly larger -- 16 GB memory  for VI (which is  on the top of capability of present GPUs)  and only 1 GB for UR, see Fig.  \ref{fig:act_per_mvmc}.

\begin{figure}
   \centering
   \includegraphics[width=14cm]{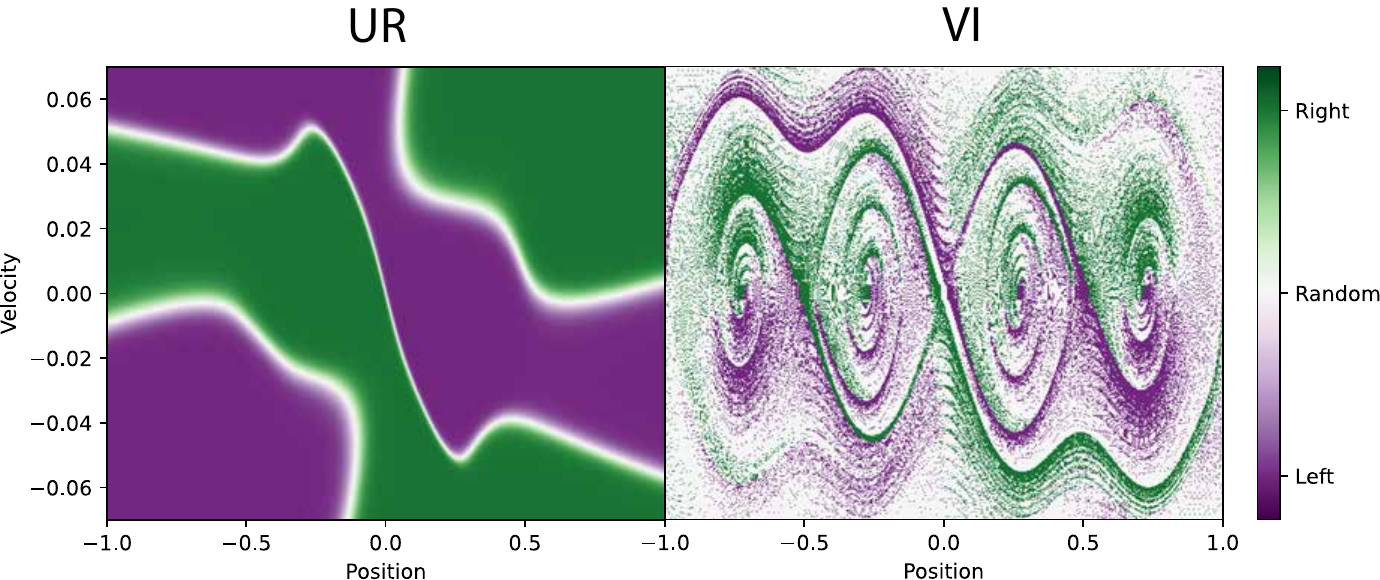}
   \caption{The color map of the final action policy for different algorithms for Multi-Valley Mountain Car problem. Left panel -- UR, right panel -- VI.}
 \label{fig:pi}
 \end{figure}

The final policies for UR and VI algorithms are depicted in Figure \ref{fig:pi} as a colour map of an action ("acceleration" $a$) as the function of a state (position $x$ and velocity $\dot{x}$). The policy of the UR manifests as four continuous regions, demonstrating symmetry of the actions with respect to the center of the system, $x=0$; the actions are natural and follow common sense. That is, it acts in order to increase the amplitude of oscillations, before reaching the destination area and to balance on the top of the hill within this area.  The policy of the VI algorithm has similar features, although it reflects the discrete nature of the algorithm. It contains a dominating area of random strategies, which is not typical for UR. This is expected for the  VI due to its discrete nature and manifests its inability to distinguish between the environment states for hard problems. Random strategy is practically lacking for the novel UR algorithm, that is, it can always discriminate between good and bad action. The abundance of random actions for VI illustrates again the superiority of UR over VI. The policy of the failed algorithms, such as PPO and UR-NE, is not shown. We just briefly mention here that UR-NE demonstrates reasonable actions in states, close to the ones with the reward and PPO algorithm behaves there randomly.

\subsection{Umbrella RL for StandUp problem}
\begin{figure}[t]
\centering
\includegraphics[width=0.9\columnwidth]{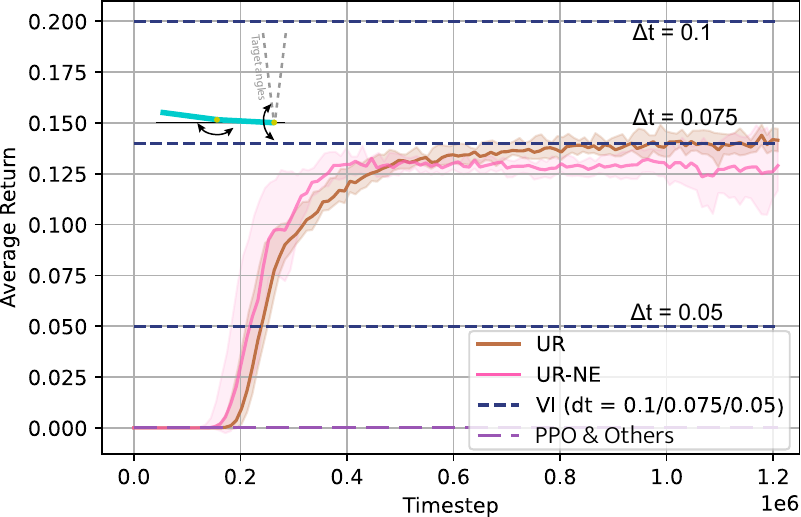}
\caption{Average expected return for StandUp problem: For UR, UR-NE, PPO, RND, and iLQR the simulation time step is  \(dt = 0.05 \) in each training run of 50 episodes (10 runs in total). For VI algorithm additional time steps of $0.075$, $0.05$ and $0.01$ are included. The total simulation time is 200. The others algorithms of the figure --  RND, iLQR  (dt = 0.05) and VI (dt = 0.01) obviosly fail.}
\label{fig:act_per_su}
\end{figure}

 The the StandUp problem is a prominent  example of the  system,  where the boundary conditions play an important role, which complicates the solution. Below we illustrate the application of the new algorithm to this hard problem.    The optimal strategy, found by the VI algorithm, reveals that it  always resides  on the boundary.  The UR algorithm is also capable to deal with such (hard) boundary conditions, see Fig. \ref{fig:act_per_su}. 

The figure also clearly demonstrates the convergence and  good performance of the UR. As it has been already demonstrated for the Multi-Value MC problem, the VI algorithm is very sensitive to the discretization time step $dt$ -- it rapidly loses its efficiency with the decreasing $dt$. While relatively large $dt$ are admissible  for weak gravity, where VI is efficient,  stronger gravity requires smaller $dt$, thus, reducing the VI power. This is illustrated in Fig. \ref{fig:act_per_su}: for large $dt=0.1$ (admissible for very small gravity) VI outperforms UR, for somewhat smaller values, $dt=0.075$ their performance is  comparable, and for still smaller $dt=0.05$, UR significantly outperforms VI. The further decreasing $dt$  does not affect the efficiency of UR, but drops the efficiency of VI to zero. Trajectory-based PPO, RND and iLQR algorithm fails similarly, which proves again, that  these widely-used  algorithms are completely useless for the hard problems. For the obvious reason (a lack of universality) we did not use here reward reshaping techniques, see the discussion below. 

\section{Discussion and conclusion}
\label{sec:con}
We propose a novel algorithm for Reinforcement Learning (RL) to solve hard problems. These problems, in contrast to regular ones, are associated with the following complications: (i) a significantly delayed  reward, given at  final states only, (ii) the existence of state traps, that  are difficult to avoid and (iii) the lack of a certain terminal state.

We demonstrate that the new algorithm being universal in implementation significantly outperforms, the state-of-the-art RL methods, most of which fail in solving hard problems. The computational superiority of the new approach will further increase with the increasing complexity of the system. The main idea of the method is to apply  a continuous (i.e., effectively infinite) ensemble of acting agents, instead of a limited number, as  in conventional RL. As the result, the modified expected return, quantifying the efficacy of the action policy, is estimated for the complete ensemble of states and actions. It comprises the conventional expected return of RL, as well as the term, specifying the ensemble entropy. To find the optimal policy we apply the appropriate modification of the policy gradient method. The policy optimization in the modified approach is entangled with the optimization of the distribution function specifying the agents' states. In  practical implementation of the method, neural networks are exploited, which makes the algorithm very efficient. 

We demonstrate the application of Umbrella RL to the Mountain Car problem  in its multi-valley  version and to the StandUp problem. Here we address the most simple models to illustrate the main ideas  of Umbrella RL for hard problems, with minimum details. 
Certainly, the new method may be applied for more practical cases, e.g.  for "Robotic arm" problem -- our StandUp problem is the according "minimalistic" model.   The Multi-Valley problem, the more complicated version of the original (single-valley), may be considered as a 'minimalistic' model of a self-driving car; it  possesses all the three difficulties of the hard problems.  

The performance of the Umbrella RL (UR) has been compared for these models with other state-of-the-art RL methods,  which are expected to be the most efficient for hard problems. Among these are the proximal policy optimization (PPO), which is believed to be the most effective and universal RL method, the  RND method, that is a prominent  method for hard exploration problems and iLQR algorithm, which  is another  popular model-based RL algorithm.  Additionally, we apply the  value iteration (VI) algorithm, which guarantees the convergence to the global optimum.
We demonstrate that the new Umbrella RL significantly outperforms all the above algorithms. Moreover,  most of these  algorithms (PPO, RND, iLQR) fail to solve the Multi-Valley Mountain Car and StandUP problem in general case,  that is, for arbitrary gravity strength. Although VI could solve these problems, it demonstrated significantly lower efficiency for sufficiently small time-step, associated with large gravity, furthermore, it required significantly larger memory. 

To illustrate the universality of Umbrella RL, we present  here its continuous-time version. Naturally, it may be also applied for any problem with discrete time. The discrete-time version may be formulated in a very similar way. This is important, since the most of RL problems are formulated in discrete time settings. 
The discrete-time  modification is based on the same ideas. Namely, one needs to assess the distribution of agents in the environment, which stems from the action-policy distribution. Then the ensemble of agents is to be used to improve the action policy. Simultaneously, it should be used to control the exploration-exploitation balance of the policy. Usually, discrete time simulations are computationally simpler. 

In the present study  we report one of the possible versions of the Umbrella RL --- the model-based version. Its generalization, for a  model-free version may be also possible and will be presented elsewhere.

\section{Acknowledgements}

EN is supported by the Analytical center under the RF Government (subsidy agreement 000000D730321P5Q0002, Grant No. 70-2021-00145 02.11.2021).

\vskip 0.2in

\bibliographystyle{elsarticle-num} 
\bibliography{main}

\appendix

\section{}

\subsection{Derivation of the evolution equation for $p(s, t)$}
\label{sec:ap_st}
The derivation of the equation for $p(s,t)$ is straightforward, as it is essentially a continuity equation. Let $p(s,t)$ be the state distribution for an ensemble of agents at time $t$. Then $p(s,t) ds$, gives the probability that an agent resides at time $t$ in a close vicinity of the (multi-component) point $s$ of the state space, where  $ds$ is the respective volume of an infinitesimal element, centered at the point $s$. Consider a closed domain $\Omega$ in the state space with the boundary  $\Gamma$, and with the vector $e(s)$ being a unit external normal to $\Gamma$ at the point $s$.  The probability current through the surface element $dS_s$, located at  the point $s$ of the boundary $\Gamma$, may be written as $p(s,t) (v\cdot e) dS_s $, where $v=v(a,s)=ds/dt$ is  the vector of local probability current. It depends on both -- the state $s$ and the action $a$, determined by the policy $\pi(a|s)$. The total local probability current equals the sum of all currents corresponding to different actions $a$, that is,  to $p(s,t) \left< v\cdot e \right>_{\pi} dS_s $, where $\left< \ldots \right>_{\pi}$ denotes the averaging over the policy $\pi$. Hence, the rate of change of the total probability to reside within the domain $\Omega$ reads, 
$$
\frac{d}{dt} \int_{\Omega} p(s,t) ds = -\int_{\Gamma} p(s,t) \left< v\cdot e \right>_{\pi} dS_s = -  \int_{\Omega}  \nabla_s \left[  p(s,t)  \left< v(a,s)\right>_{\pi} \right]ds, 
$$
where Gauss theorem has been applied. Since the domain $\Omega$ is arbitrary, we arrive at the continuity equation, 
\begin{equation}
    \label{4a}
 \frac{dp}{dt} + \nabla_s  p  \left< v(a,s)\right>_{\pi}   =0, 
\end{equation}
which is Eq. \eqref{4} of the main text.

\subsection{Differential policy gradient}
\label{sec:ap_dif_pg}
To remind the main notations and basic ideas of the policy gradient derivation we start from the case of discrete time and skip the dependence on the entropy, which we add later. That is, we start from the expected reward  \(J(\pi) \) of the form, 
\begin{equation}
   J(\pi) =  \E_{\substack{ s_t \sim p_{a,t} \\ a_t \sim \pi }}\left[\sum_{t=0}^\infty \gamma^{t} r(s_t,a_t)\right]. 
\end{equation}
Here we abbreviate, $s_t=s(t)$, $a_t=a(t)$, $p_{a,0}=p(s,0)=p_0(s)$ and $p_{a,t+1}=p(s_{t+1}|s_{t}, a_{t})$ for $t\geq 1$. In our case $p(s_{t+1}|s_{t}, a_{t})$ is a predefined function, that is,  $s_{t+1} = s_{t} +v(s_{t},a_{t})$.  These functions are respectively, the state, action and distribution of the ensemble states, for discrete time instants $t=0,1, \ldots , \infty$. Here we apply the standard notations, which imply that the averaging is to be performed  over the state distribution $p_{a,t}$  and  over the policy $\pi_{\theta} (a|s)$, parametrized by the (multi-component) parameter $\theta$. Note that while $p_0(s)$ is the initial state distribution, all other functions $p_{a,t}$ are determined by the previous state and action.  Application of the standard derivation (see e.g. \cite{sutton1999policy}), step by step,  yields the gradient of $ J(\pi)$ with respect to  $\theta$:

\begin{equation}
\label{DPG1}
    \grad_{\theta} J(\pi_{\theta}) = 
    \E_{\substack{ s_t \sim p_{a,t} \\ a_t \sim \pi }}     \left[\sum_{t=0}^{\infty}  \gamma^{t} \grad_\theta \ln \pi_{\theta}(a_t|s_t) \left\{ 
    \sum_{k=t}^{\infty} \gamma^{k-t} r(s_k,a_k) \right\} \right] .
\end{equation}
The sum in the curly brackets may be written as $\sum_{k=t}^{\infty} \gamma^{k-t} r(s_k,a_k) =r(s_t,a_t) + \sum_{k=t+1}^{\infty} \gamma^{k-t} r(s_k,a_k)$, where $s_{t+1}$ (for $k=t+1$) depends on $s_t$ and $a_t$, that is $s_{t+1} =s_t+v(s_t,a_t)$. This motivates us to introduce the function $Q(s,a)$,
\begin{equation}
\label{DPG2a}
Q(s,a) = r(s,a) +\gamma \E_{\substack{ s_t \sim p_{a,t} \\ a_t \sim \pi }}     \left[  
    \sum_{i=1}^{\infty} \gamma^{i-1} r(s_i,a_i) \Big|_{ s_0=s,a_0=a}\right] , 
\end{equation}
which is termed in RL as "action-value function". Then Eq. \eqref{DPG1} takes the form, 
\begin{equation}
\label{DPG1d}
    \grad_{\theta} J(\pi_{\theta}) = 
    \E_{\substack{ s_t \sim p_{a,t} \\ a_t \sim \pi }}     \left[\sum_{t=0}^{\infty}  \gamma^{t} \grad_\theta \ln \pi_{\theta}(a_t|s_t) Q(s_t,a_t) \right]
\end{equation}
Averaging $Q(s,a)$ w.r.t. the actions with the policy $\pi$, we obtain another function, 
\begin{equation}
\label{DPG2b}
V(s) =  \E_{\substack{ s_t \sim p_{a,t} \\ a_t \sim \pi }}     \left[  
    \sum_{i=0}^{\infty} \gamma^{i} r(s_i,a_i) \Big|_{ s_0=s} \right] , 
\end{equation}
termed in RL as "value function"; note that it depends only on state.  The  action-value function may be then written in the form, .
\begin{equation}
\label{DPG2c}
Q(s,a) =r(s,a) +\gamma V\big(s+v(s,a) \big).
\end{equation}

Using the fact that for any arbitrary state function $F(s_t)$,
\begin{equation}
\label{Ezero} \E_{\substack{ s_t \sim p_{a,t} \\ a_t \sim \pi }}  \left[ \grad_\theta \ln \pi_{\theta}(a_t|s_t) F(s_t) \right]=
\E_{ s_t \sim p_{a,t} } F(s_t) \grad_\theta \left\{ \int \pi_{\theta}(a_t|s_t) da_t \right\} =0, 
\end{equation}
since $\int \pi_{\theta}(a_t|s_t) da_t = 1$, we can subtract in Eq. \eqref{DPG1d} $V(s_t)$ from $Q(s_t,a_t)$. That is, we can write:  
\begin{equation}
\label{DPG1d2}
    \grad_{\theta} J(\pi_{\theta}) = 
    \E_{\substack{ s_t \sim p_{a,t} \\ a_t \sim \pi }}     \left[\sum_{t=0}^{\infty}  \gamma^{t} \grad_\theta \ln \pi_{\theta}(a_t|s_t) A(s_t,a_t) \right], 
\end{equation}
where we introduce the function, $A(s,a)$, known  as "advantage" \cite{sutton1999policy}, 
\begin{equation}
    \label{Adef}
A(s,a) =Q(s,a)-V(s,a). 
\end{equation}
Utilizing $A(s_t,a_t)$, instead of $Q(s_t,a_t)$,  results in a more smooth behavior of $\grad_{\theta} J(\pi_{\theta})$, when it is estimated using a limited series of states and actions, $\left\{s_t,a_t \right\}$  \cite{sutton1999policy}. 
Now we recast Eq. \eqref{DPG1d2} into the form
\begin{equation}
\label{DPG1e}
\begin{split}
    \grad_{\theta} J(\pi_{\theta}) &=  \E_{a \sim \pi }\left[ \left( \sum_{t=0}^{\infty}  \gamma^{t} p(s,t) \right) 
     \grad_\theta \ln \pi_{\theta}(a|s) A(s,a) 
     \right] \\
     &= \E_{\substack{ s \sim \overline{p} \\ a\sim \pi }} \left[  \grad_\theta \ln \pi_{\theta}(a|s) A(s,a) \right],
\end{split}
\end{equation}
Where $p(s,t)$ is the state distribution at time-step $t$ averaged over initial distribution $p_0(s)$ and all actions and states at the previous steps, up to $t-1$:   
$$
p(s,t)=\E_{\substack{ s_0 \ldots s_{t-1} \sim p_{a,0} \ldots p_{a,t-1} \\ a_0, \ldots a_{t-1} \sim \pi }} p_{a,t} = 
\E_{\substack{  s_0 \ldots s_{t-1} \sim p_{a,0} \ldots p_{a,t-1}  \\ a_0, \ldots a_{t-1} \sim \pi} } p(s_t|s_{t-1},a_{t-1})
$$
and we introduce the averaged distribution $\overline{p}(s)$:
\begin{equation}
\label{DPG1g}
\overline{p}(s) = \sum_{t=0}^{\infty}  \gamma^{t} p(s_t). 
\end{equation}

Consider now the continuous-time  counterpart of Eq. \eqref{DPG1e}. The summation is to be transformed into integration, yielding for the averages distribution, 
\begin{equation}
\label{pcont1}
\overline{p}(s) = \int_0^\infty \gamma^t p(s,t) dt,  
\end{equation}
which coincides with the definition \eqref{pav} of the main text; the normalizing factor $| \ln \gamma |$ is also omitted. As the state varies continuously, with the rate, $\dot{s} = v(s,s)$, the increment of a state is proportional to the infinitesimal time steps, $dt$, that is, $ds =v dt$. Correspondingly, the reward transforms into the reward rate $r(s,a)$, resulting in the reward $r(s,a)dt$ accumulated during time interval $dt$. The advantage function,  introduced in Eq. \eqref{Adef}, also transforms into the respective rate,  quantifying the advantage, $A(s,a)dt$ accumulated  during a time interval $dt$. That is, the continuous counterpart of Eq. \eqref{Adef} reads,  
\begin{equation}
\label{Acont}
A(s,a)dt = Q(s,a)-V(s).
\end{equation}
The discount factor $\gamma$, weighting the rewards of  the successive states, transforms now into the factor $\gamma^{dt}$ -- the discount factor for two time instants, separated by the infinitesimal time interval $dt$. This yields the continuous counterpart of Eq. \eqref{DPG2c}: 
\begin{equation}
\label{Qcont}
    Q(s,a) = \gamma^{dt} V\big(s + v(s,a)dt\big) + r(s,a) dt, 
\end{equation}
indicating that for $dt \to 0$ the function $Q(s,a)$ converts into the value function $V(s)$, as the reward vanishes in this limit.
Using Eqs. \eqref{Acont} and \eqref{Qcont} we obtain, keeping only linear -order terms of $dt$ for the function $Q(s,a)$, 
\begin{equation}
\label{AQV}
\begin{split}
A(s,a)dt &= (1+ \log \gamma \, dt +\ldots )\left( V(s) + v(s,a) \cdot \nabla_s V(s) dt +\ldots \right) dt\\
&+ r(s,a) dt -V(s). 
\end{split}
\end{equation}
Dividing the above equation by $dt$,  we find, after some algebra, in the limit $dt \to 0$:
\begin{equation}
\label{Afin}
    A(s,a) = r(s,a)+ v(s,a) \cdot \nabla_s V(s)   + \log {\gamma} \, V(s)
\end{equation}
With the new meaning of the function $A(s,a)$, Eq. \eqref{Afin}, and the averaged distribution, Eq. \eqref{pcont1}, the form of the policy gradient, Eq. \eqref{DPG1e} persists. 

Finally, consider the policy gradient with the entropic term, that is, we need to compute the expression (see Eq. \eqref{Hdef1} of the main text): 
\begin{equation}
\begin{split}
    \nabla_{\theta} H[\overline{p}\cdot \pi]&= - \nabla_{\theta} \left< \E_{ a\sim \pi_{\theta} }\left[ \log \overline{p} \cdot \pi_{\theta} \right]
\right>_{\overline{p} } = -  \nabla_{\theta} \E_{\substack{ s_t \sim p_{a,t} \\ a_t \sim \pi }}\left[ \sum_{t=0}^{\infty} \gamma^t \left[\log \overline{p} \cdot \pi_{\theta}\right] \right]  \\
&= -\E_{\substack{ s_t \sim p_{a,t} \\ a_t \sim \pi }}     \left[\sum_{t=0}^{\infty}  \gamma^{t} \grad_\theta \ln \pi_{\theta}(a_t|s_t) \left\{ 
    \sum_{k=t}^{\infty} \gamma^{k-t} \log \overline{p}(s_k) \cdot \pi_{\theta}(a_k|s_k) \right\} \right]+\\
    &-   \E_{\substack{ s_t \sim p_{a,t} \\ a_t \sim \pi }}\left[ \sum_{t=0}^{\infty} \gamma^t  \nabla_{\theta} \left[\log \overline{p} \cdot \pi_{\theta}\right] \right],
\end{split}
\end{equation}
where we considered that the probabilities under the logarithm depend on policy parameters \(\theta\). In that case

\begin{eqnarray}
    \E_{\substack{ s_t \sim p_{a,t} \\ a_t \sim \pi }}\left[ \sum_{t=0}^{\infty} \gamma^t  \nabla_{\theta} \left[\log \overline{p} \cdot \pi_{\theta}\right] \right] = -\nabla_{\theta} \left< \E_{ a\sim \pi_{\theta} }\left[ \nabla_{\theta} \log \overline{p} \cdot \pi_{\theta} \right]  
\right>_{\overline{p} } = 0
\end{eqnarray}

Since \(\int \left[\overline{p}\cdot\pi \right] dsda = 1\). And the first term 
\begin{equation}
\begin{split}
    &-\E_{\substack{ s_t \sim p_{a,t} \\ a_t \sim \pi }}     \left[\sum_{t=0}^{\infty}  \gamma^{t} \grad_\theta \ln \pi_{\theta}(a_t|s_t) \left\{ 
    \sum_{k=t}^{\infty} \gamma^{k-t} \log \overline{p}(s_k) \cdot \pi_{\theta}(a_k|s_k) \right\} \right] \\
    &=     \E_{\substack{ s_t \sim p_{a,t} \\ a_t \sim \pi }}     \left[\sum_{t=0}^{\infty}  \gamma^{t} \grad_\theta \ln \pi_{\theta}(a_t|s_t) \left\{ 
    \sum_{k=t}^{\infty} \gamma^{k-t} (-\log \overline{p} \cdot \pi )\right\} \right]
\end{split}
\end{equation}
is responsible for the reward associated with the entropy, which is \( -\log \overline{p} \cdot \pi\) in the equation.  Taking into account the above derivation we obtain
\begin{equation}
\begin{split}
\label{DelH}
    \nabla_{\theta} H[\overline{p}\cdot \pi]&= \E_{\substack{ s_t \sim p_{a,t} \\ a_t \sim \pi }}     \left[\sum_{t=0}^{\infty}  \gamma^{t} \grad_\theta \ln \pi_{\theta}(a_t|s_t) \left\{ 
    \sum_{k=t}^{\infty} \gamma^{k-t} (-\log \overline{p} \cdot \pi ) \right\} \right],
\end{split}
\end{equation}
that means, that the entropy term inclusion is equivalent to an extra reward term addition.

Combining Eq. \eqref{DPG1} and \eqref{DelH}, we obtain for the policy gradient for the umbrella RL expected return, ${\cal J}^{\rm URL} = J(\pi) +\alpha H[p\cdot \pi]$:
\begin{equation}
\label{8a}
     \grad_\theta {\cal J}^{\rm URL}(\pi_\theta) =  \left<  \E_{a \sim \pi } \big[\grad_\theta \log  \pi_\theta(a|s) \,  A_{\rm u}(s,a)\big]  \right>_{\overline{p}},  
\end{equation}
where the Umbrella RL differential advantage function, $A_{\rm u}(s,a)$,  results from $A(s,a)$ defined in Eq. \eqref{Afin} and from the last expression in Eq. \eqref{DelH}:
\begin{equation}
\label{AfinURL}
    A_{\rm u}(s,a) = r(s,a)+ v(s,a) \cdot \nabla_s V(s)   + \log {\gamma} \, V(s) -\tilde{\alpha} \log \left( \overline{p} \cdot \pi_{\theta} \right), 
\end{equation}
where $\tilde{\alpha} = |\log \gamma| \alpha$, that is, we restore the normalization factor $|\log \gamma|$, skipped in the derivation. 
Eqs. \eqref{8a} and \eqref{AfinURL} coincide with the according Eqs. \eqref{8} and \eqref{9} of the main text. Note that the structure of the differential advantage function persist -- one only needs to use renormalized reward: $r_{\rm u}(s,a)=r(s,a) - \tilde{\alpha} \log \left( \overline{p} \cdot \pi_{\theta} \right)$.

\subsection{Derivation of the steady-state equations for the value function $V_{\rm u}(s)$ }
\label{sec:ap_value} 

The continuum-time counterpart of the value function reads, 
\begin{equation}
\label{Vcont}
    V_{\rm u}(s,t) = \int_0^{\infty} \gamma^t \E_{a \sim \pi(a|s)} \bigg[r_{\rm u}(s,a) \bigg] dt .  
\end{equation}
As for the case of average distribution, we also define "incomplete" value function for Umbrella RL, 
\begin{equation}
\label{Vst}
    V_{\rm u}(s,t) = \int_0^{t} \gamma^t \E_{a \sim \pi(a|s)} \bigg[r_{\rm u}(s,a) \bigg] dt , 
\end{equation}
where $r_{\rm u}(s,a)=r(s,a)- \tilde{\alpha} \,\log( \overline{p} \pi )$ is the effective reward for the Umbrella RL, which accounts for the distribution entropy [see Eq. \eqref{Hdef1} of the main text]. The respective value function, defined by Eq. \eqref{10} of the main text, results from the "incomplete" value function for $ t\to \infty$, that is, $V_{\rm u}(s)=V_{\rm u}(s, \infty)$. The Bellman Equation \cite{bellman1957markovian} applied to $V_{\rm u}(s,t)$ in Eq. \eqref{Vst} yields (see also \cite{howard1960dynamic}):  
\begin{equation}
\label{Vst1}
    V_{\rm u}(s,t) = \E_{a \sim \pi(a|s)} \bigg[ \gamma^{dt} V_{\rm u}\big(s + v(s,a)dt, t-dt \big) + r_{\rm u}(s,a) dt \bigg],
\end{equation}
Expanding $V_{\rm u}\big(s + v(s,a)dt, t-dt \big)$ in Eq. \eqref{Vst1}, up to linear terms in $dt$, yields, 
\begin{equation}
\begin{split}
    V_{\rm u} (s,t) &\simeq   V_{\rm u}(s,t) + \\
    &+ \bigg[ \log \, \gamma  V_{\rm u}(s,t) +\E_{a \sim \pi(a|s)} \left[v(s,a) \cdot \nabla_s V_{\rm u}(s,t) + r_{\rm u}(s,a)\right] -  \dot{V}_{\rm u}(s,t) \bigg] dt.
\end{split}
\end{equation}
Hence, in continuous case the incomplete value function obeys the equation, 

\begin{equation}
\label{eq:dve_de}
    \dot{V}_{\rm u}(s,t) = \E_{a \sim \pi(a|s)} A_{\rm u}(s,a,t),
\end{equation}
where
\begin{equation}
    A_{\rm u}(s,a,t) = \log  \gamma \, V_{\rm u}(s,t) +v(s,a) \cdot \nabla_s V_{\rm u}(s,t) + r_{\rm u}(s,a). 
\end{equation}
In the limit $t \to \infty$, $\dot{V}_{\rm u}(s,t)=0$, as it follows from its definition \eqref{Vst},  $V_{\rm u}(s)=V_{\rm u}(s, \infty)$, and we obtain, 
\begin{equation}
    \E_{a \sim \pi(a|s)} A_{\rm u}(s,a) = \overline{v}  \cdot \nabla_s V_{\rm u}(s) + \overline{r}_u - |\log \gamma |\, V_{\rm u}(s) =0 , 
    \end{equation}
which corresponds to Eq. \eqref{9} of the main text. Here, as defined previously, $\overline{v} =  \E_{a \sim \pi} \left[v (s,a) \right] $ and  $\overline{r}_u =  \E_{a \sim \pi} \left[r_u (s,a) \right]$. We also define the function 
\begin{equation}
\begin{split}
\label{As}
    A_{\rm u}(s,a)&=A_{\rm u}(s,a, \infty) \\
    &= r(s,a) - \tilde{\alpha} \,  \log \big(\overline{p} \cdot \pi\big)+ v \cdot \nabla_s V_{\rm u}(s)   - |\log {\gamma} | \,  V_{\rm u}(s), 
\end{split}
\end{equation}
which is the same definition as given by Eq. \eqref{9} of the main text.  It may be shown that the function  $A_{\rm u}(s,a)$ is the continuous counterpart of the one of the basic RL functions -- the advantage function (see Appendix (\ref{sec:ap_dif_pg})).

\subsection{Derivation of the steady-state equation for the averaged distribution $\overline{p}(s)$}
\label{sec:ap_ss_for_p}
First, we introduce the incomplete averaged distribution:
\begin{equation}
    \label{pst}
\overline{p}(s,t) =c  \int_0^t \gamma^{t'} p(s,t') dt', 
\end{equation}
where $c=c(t)=\log \gamma (\gamma^t-1)^{-1}$ is the normalization constant, so that $\int \overline{p}(s,t) ds=1$ and $c(\infty)= -\log \gamma$. The averaged distribution $\overline{p}(s)$, defined by Eq. \eqref{4} of the main text may be expressed as $\overline{p}(s)= \overline{p}(s,\infty)$. Then from Eq. \eqref{pst}  follows, 

\begin{equation}
\label{ntt}
\begin{split}
    \overline{p}(s,{t+dt}) &= c(t+dt)\int_0^{t+dt} \gamma^{t'} p(s,t')d't \\
    &= \left(c(t)+c'(t) dt \right) \left[\gamma^{dt} \int_0^{t} \gamma^{t_1} p(s,t_1+dt)dt_1 + p_0(s) dt\right],
\end{split}
\end{equation}
where we keep only linear-order terms with respect to the small time increment $dt$ and $p_0(s)=p(s,0)$ is the initial distribution. Using 
$$
p(s,t_1+dt)= p(s,t_1)-\nabla_s \left[p(s,t_1)  \left< v(a,s)\right>_{\pi} \right] dt, 
$$ 
which follows from the evolution equation \eqref{4} of the main text, we recast \eqref{ntt} into the form, 

\begin{equation}
\label{ptt}
\begin{split}
        \overline{p}(s,{t+dt}) &= \left(c(t)+c'(t) dt \right) \gamma^{dt} \int_0^{t} \gamma^{t_1} \left[p(s,t_1) - \nabla_s p(s,t_1)  \left< v(a,s)\right>_{\pi} dt \right] dt_1 + \\
        &\ + c(t)p_0(s) dt \\
        &=  \left(c(t)+c'(t) dt \right) \gamma^{dt} \, \frac{\overline{p}(s,t)}{c(t)}+  \\
        &-\left(c(t)+c'(t) dt \right) dt \,\int_0^{t} \gamma^{t_1} \nabla_s \left[p(s,t_1)  \left< v(a,s)\right>_{\pi} \right] dt_1+ c(t) p_0(s) dt \\
        &= \overline{p}(s,t) + c'(t)\frac{\overline{p}(s,t)}{c(t)}  dt +\\
        &+\log \gamma  \,\overline{p}(s,t) dt - c(t) \, \nabla_s  \left[ \frac{\overline{p}(s,t)}{c(t)} \left< v\right>_{\pi} \right] dt + c(t) p_0(s) dt \\ 
        &= \overline{p}(s,t) +\\
        &+\left[  \log \gamma \, \overline{p}(s,t) +\frac{ c'(t)}{c(t)}\overline{p}(s,t) -\nabla_s   \overline{p}(s,t) \left< v\right>_{\pi}  + c(t) p_0(s) \right] dt , 
\end{split}
\end{equation}
where we again keep only linear terms in $dt$ take into account that $\left< v\right>_{\pi}$ does not depend on $t_1$. Dividing both sides of Eq. \eqref{ptt} by $dt$, we obtain in the limit $dt \to 0$:
\begin{equation}
    \label{ptt1}
    \frac{d}{dt}  \overline{p}(s,t) = \log \gamma \, \overline{p}(s,t) +\frac{c'(t)}{c(t)}\overline{p}(s,t) - \nabla_s   \overline{p}(s,t) \left< v\right>_{\pi}  + c(t) p_0(s).
\end{equation}
The above equation for the incomplete average distribution $\overline{p}(s,t)$, may be used to obtain the equation for $\overline{p}(s)$, by taking the limit, $t \to \infty$. It directly follows from the definition \eqref{pst}, that $d \overline{p}(s,t)/dt \to 0$ for $t \to \infty$ (recall that $\gamma <1$); similarly $c'(\infty)=0$ and $c(\infty)= -\log \gamma$. Hence,  in the $t \to \infty$ limit, we obtain from Eq. \eqref{ptt1}:
\begin{equation}
    \label{ptt2}
     \log \gamma \,\left( \overline{p}(s)  - p_0(s) \right) - \nabla_s   \overline{p}(s) \left< v\right>_{\pi}=G^{\pi}(s)=0, 
\end{equation}
which is Eq. \eqref{14} of the main text.

\subsection{Approximate steady state}
\label{sec:ap_ass}
In this section we consider a simplified, heuristic, approach to find an approximate steady state. A more rigorous, although more technical derivation is presented in the next section. 

It was shown in Sec \ref{sec:ap_value}, that the value function is a steady-state solution for the evolution equation of an "incomplete" value function. This property of the function suggests the application of the standard technique of gradient descent approach. To this end we define the loss-function:

\begin{equation}
Loss(V) = \E_{ s \sim \rho}\left[\overline{v}\cdot \nabla_s V_u(s)  + \overline{r}_u - |\log {\gamma} | \,  V_u(s)\right]^2,
\end{equation}
where \(\rho\) is an arbitrary sampling distribution, which covers the observation range. The minimum of the loss function,  w.r.t. the value function,  is satisfied for the steady value of this function. Hence, the  gradient of the loss-function, parametrized with $\phi$, may be calculated as follows:

\begin{equation}
\begin{split}
    \grad_\phi Loss(V)&=2  \E_{ s \sim \rho}\left[\overline{v}\cdot \nabla_s V(s)  + \overline{r} - |\log {\gamma} | \,  V(s)\right] \cdot\\
    &\cdot  \grad_\phi \left[\overline{v}\cdot \nabla_s V(s)  + \overline{r} - |\log {\gamma} | \,  V(s) \right].
\end{split}
\end{equation}
Usually, the estimation of the second-order gradients  is computationally costly, moreover,  they may generate additional local minima. So, in the present work, we assume that the second-order gradients may be neglected and that it is sufficient to account only for the first-order gradients. This approximation yields, 

\begin{equation}
\begin{split}
        \grad_\phi Loss(V)  &\approx -2 |\log {\gamma} |  \E_{ s \sim \rho} \grad_\phi  \,  V(s) \left[\overline{v}\cdot \nabla_s V(s)  + \overline{r} - |\log {\gamma} | \,  V(s)\right], 
\end{split}
\end{equation}
which may be rewritten, for simplicity, as
\begin{equation}
    \grad_\phi Loss(V)  = - \left< \E_{a \sim \pi} \left[\grad_{\phi}{V (s)} \cdot A(s,a) \right]\right>_{\rho},
\end{equation}
where we omit the factor \(2 |\log \gamma|\). It is not important, since in the practical implementation of the gradient decent, the gradient is, anyway, scaled,  with a tunable learning rate factor, see below. Therefore we obtain,  

\begin{equation}
\label{eq:d_value}
    {\bf \phi}_{k+1} =\phi_k + \beta^{V} \left< \E_{a \sim \pi} \left[\grad_{\phi }{V (s)} \cdot A(s,a) \right]\right>_{\rho}, 
\end{equation}
which is Eq. \eqref{phik}  of the main text.  Similarly, we can estimate the steady state solution of average agents distribution \(\overline{p}\). In that case, the deviation from its steady state is the following:

\begin{equation}
Loss(\overline{p}(s)) = \left[ \nabla_s\cdot \overline{p}(s) \overline{v} -\log \gamma \, \left(\overline{p}(s) -p_0(s) \right)\right]^2.
\end{equation}
To find the gradient estimate we omit  the  divergence  term, \( \nabla_s\cdot \overline{p}(s) \overline{v}\), of the loss function, parameterized by $\eta$, which results in the following approximation: 

\begin{equation}
\begin{split}
    \grad_\eta Loss(\overline{p}(s)) &\approx -2 \log \gamma \cdot \grad_\eta \overline{p}(s) \cdot G(\overline{p},s)\\
    &=  -2 \log \gamma \cdot \overline{p}(s) \cdot \grad_\eta  \log \overline{p}(s) \cdot G(\overline{p}(s),s).
\end{split}
\end{equation}
We wish to average the gradient of the loss function over states \(s\) and simultaneously recast it into the form of Eqs. \eqref{8}, \eqref{11} of the main text. This may be done choosing the distribution \(P(s) = C \cdot \rho(s)/\overline{p}(s)\), where \(C\) is the normalization factor. Next, averaging the gradient with state weights \(\rho/\overline{p}\) we receive a full state gradient estimate

\begin{eqnarray}
    \grad_\eta Loss(\overline{p}) =- 2 |\log \gamma| C \cdot \left< \grad_\eta \log \overline{p}(s) \cdot G(\overline{p}(s),s) \right>_{\rho}.
\end{eqnarray}
Omitting the factor \(2 |\log \gamma| C \) we arrive finally at,  

\begin{equation}
\label{77}
    {\bf \eta}_{k+1} =\eta_k + \beta^{P} \left< \grad_\eta \log \overline{p}(s) \cdot G(\overline{p}(s),s) \right>_{\rho}, 
\end{equation}
which is Eq. \eqref{etak} of the main text. 

\subsection{State rate divergence}
\label{sec:ap_state_div}

State rate divergence is required for commutation of average agents distribution \(\overline{p}\) (see Eq. \eqref{14a} of the main text). It came from the divergence term \(\nabla_s\cdot \overline{p}(s) \overline{v}\). Since we suppose using a stochastic estimate of average velocity \(\overline{v}\), the proper divergence is essential. In the present work, we propose to expand the divergence as follows:

\begin{equation}
\label{108}
    \nabla_s\cdot \overline{p}  \overline{v} =  \overline{p} \E_{a \sim \pi}\left[\nabla_s\cdot v + v \cdot \grad_s{\ln \left(\pi \cdot \overline{p}\right)}\right],
\end{equation}

where \(\nabla_s\cdot v\) is state rate divergence, which is a derivative of the environment. The equation above allows a stochastic estimate of the divergence term, instead of exact computation.

Next, we show the derivation of the state rate divergence for Multi-Valley MC and StandUp environments. 

In Multi-Valleys Mountain Car problem,  the rate of change of agent state is the following (see Eq. \eqref{sdotv} of the main text):
\begin{equation}
    v(s,a)=v(x,\dot{x},a) =
    \begin{pmatrix}
        \dot{x} \\
         \left(2 \cdot a - 1\right) \cdot f - y'(x) \cdot g.
    \end{pmatrix}
\end{equation}
Note that in this specific case, the term \(\nabla_s\cdot v\) in Eq. \eqref{108} vanishes: 
\begin{equation}
\label{110}
   \nabla_s \cdot v = {\frac{\partial \dot{x}}{\partial x}} + {\frac{\partial}{\partial \dot{x}}\Big[\left(2 \cdot a - 1\right) \cdot f - y'(x) \cdot g\Big]} = 0.
\end{equation}

Similarly, according to the definition in the description of StandUp problem (see Eqs. \eqref{phi1},\eqref{phi2} of the main text), we have the following state rates:

\begin{eqnarray}
    v(s,a)= v(\varphi_1,\varphi_2,a)= 
    \begin{pmatrix}
        m_1-m_2 - g \cos \varphi_1\\ 
        m_2 - g \cos (\varphi_1+\varphi_2)\
    \end{pmatrix}.
\end{eqnarray}

Consequently, the state rate divergence, the term \(\nabla_s\cdot v\) in Eq. \eqref{108}:

\begin{equation}
\label{112}
   \nabla_s \cdot v = \frac{\partial \dot{\varphi}_1}{\partial \varphi_1} + \frac{\partial \dot{\varphi}_2}{\partial \varphi_2} =  g \left( \sin \varphi_1 + \sin (\varphi_1+\varphi_2)\right).
\end{equation}

In the present UR algorithm we use the terms \(\grad_s \cdot v\) from Eq. \eqref{110} and \ref{112} and calculate the stochastic estimate of the divergence term \(\grad_s \cdot p \overline{v}\) for specific action samples. Specifically we substitute the avenging in Eq. \eqref{108} with averaging w.r.t. finite number of action samples.

\subsection{Boundary conditions } 
\label{sec:ap_bc_mvmc}
\subsubsection{Boundary conditions for neural networks}

The boundary conditions are a part of the environment. They are applied to the probability distribution and value function and define,  what are the properties  of the value function and agent distributions for the observation space boundary. In the present work, we embed directly the boundary conditions  in the neural network. Let \(NN(x)\) be a neural network with a scalar output and input vector \(x\); we impose von Neumann boundary conditions:

\begin{eqnarray}
   \frac{dNN(x)}{dn}\bigg|_{\Gamma} = 0,
\end{eqnarray}
where \(\Gamma\) is the boundary and  \(n\) denotes the extrnal normal to the boundary \(\Gamma\). 
In the present work, we combine the NN of two parts such as:

\begin{equation}
    NN(x) = NN_h(h(x)),
\end{equation}
where \(h\) is environment specific representation function and \(NN_h\) is a neural network taking state representation \(h\) as an input. The goal of the representation function is to guarantee  the boundary conditions. If we impose the representation \(h(x)\) as,
\begin{equation}
    \frac{d h(x)}{dx} \cdot n \bigg|_{\Gamma} = 0,
\end{equation}
then

\begin{equation}
     \frac{dNN(x)}{dn}\bigg|_{\Gamma} = \frac{dNN_h(h)}{dh}\frac{dh(x)}{dx} \cdot n \bigg|_{\Gamma} = 0.
\end{equation}
That is, So, manipulating the state representation, we can impose arbitrary(including the reflecting) boundary conditions.

\subsubsection{ Boundary conditions for Multi-Valley Mountain Car}

Reflective representation \(h\) was applied to impose the probability distribution and value function to satisfy boundary conditions (it is not required for the action policy). As a hidden state, we generate a three-dimensional vector h, such as:

\begin{equation}
    h = \begin{pmatrix}
        x\\
        \hat{v}^2\\
        d_{min} \cdot \hat{v}
    \end{pmatrix},
\end{equation}
where,

\begin{equation}
    \hat{v} = \cos\left(\frac{v-v_{min}}{v_{max} - v_{min}} \pi\right)
\end{equation}
is normalized velocity, and

\begin{equation}
    d_{min} = \min\big(|x-x_{min}|,|x-x_{max}|\big)
\end{equation}
is a distance to the boundary. Boundary values of the agent speed are \(v_{min/max} = \pm 0.07\) and boundary positions are \(x_{min/max} = \pm 0.99\).
The hidden representation was developed, such that the boundary conditions
\begin{eqnarray}
    \frac{dh}{dv}\bigg|_{v = v_{min}/v_{max}} = 0
\end{eqnarray}
and
\begin{eqnarray}
   h(x,v)\bigg|_{x = x_{min}/x_{max}} = h(x,-v)\bigg|_{x = x_{min}/x_{max}}
\end{eqnarray}
were satisfied. This results in the corresponding reflection boundary conditions for the agents probability distribution \(\overline{p}\) and the value function \(V\):

\begin{eqnarray}
    \frac{dNN(x,v)}{dv}\bigg|_{v = v_{min}/v_{max}} = 0
\end{eqnarray}
 and
\begin{eqnarray}
   NN(x,v)\bigg|_{x = x_{min}/x_{max}} = NN(x,-v)\bigg|_{x = x_{min}/x_{max}}.
\end{eqnarray}
 

\subsubsection{Boundary conditions for StandUp problem}
\label{sec:ap_bc_standup}

The boundary conditions representation for StandUp problem is defined for environment outputs \(\phi_1 \in (0,\pi)\) and \(\phi_2 \in (-\pi,\pi)\). The state representation was developed to reflect approximately  the environment dynamic on the state boundary, which is required for a proper work of the UR algorithm. The state representation translates the space with a complex shape, which has, specifically, a restriction of the form \(\phi_2 \in (-2\phi_1 , 2 \pi - 2\phi_1)\),  to a simple square domain  \( \theta_1,\overline{\theta}_2 \in (-\pi/2,\pi/2)\). The new space allows assigning of reflective boundary conditions with a simple \(\sin\) function.  That is, To be precise, the state representation reads,
\begin{equation}
    h = \begin{pmatrix}
        \sin \theta_1\\
        \sin \overline{\theta}_2
    \end{pmatrix},
\end{equation}
where

\begin{equation}
    \theta_1 = \phi_1 - \frac{\pi}{2},
\end{equation}

\begin{equation}
    \overline{\theta}_2 = \frac{1}{2}\frac{\theta_2}{1 -  |\theta_1 / \pi|},
\end{equation}
and 

\begin{equation}
    \theta_2 =  \phi_1 + \phi_2 - \frac{\pi}{2} 
\end{equation}
One can show that such state representation guarantees, that:
\begin{equation}
    \frac{dNN}{d \theta_1} \bigg|_{\theta_1 = \pm \pi/2} = 0
\end{equation}
and
\begin{equation}
    \frac{dNN}{d \overline{\theta}_2} \bigg|_{\overline{\theta}_2 = \pm \pi/2} = 0.
\end{equation}

\subsection{General approach for finding an approximate steady state 
}
\label{sec:ap_ass_g}

Assume, that we need to find approximately a steady-state solution of the following differential equation:
\begin{equation}
    \dot{y}(x) = f(x,y,\grad y),
\end{equation}
where \(x\) is arbitrary vector, \(y(x)\)  and \(f(x,y, \grad y)\) are arbitrary scalar-functions.

Here the function \(y(x)\) is specified in a class of functions, parameterized by parameters \(\theta\); we denote it by \(y_\theta (x)\) and define a finite set of function arguments \(X\) for which we will find an optimal approximation.

In that case, the primary differential equation may be rewritten as the following system of equations for each \(x \in X\):

\begin{equation}
    \dot{y} = f,
\end{equation}
where \(y = y_\theta(X) \) and \(f = f(X,y(X),\grad y(X))\).
The last equation may be written in terms of parameters \(\theta\) (here we utilize numerator layout notation):

\begin{equation}
\label{80}
    \dot{y} = \left(\grad_\theta y\right)^T \cdot \dot{\theta} = f.
\end{equation}
Next, we multiply both sides of the equation with a positive diagonal matrix \(diag(\sqrt{p})\):

\begin{equation}
\label{81}
   diag(\sqrt{p}) \cdot \left(\grad_\theta y\right)^T \cdot \dot{\theta} = diag(\sqrt{p}) \cdot f.
\end{equation}
It allows us additional degree of freedom in the the solution selection. We apply the Moor-Penrose inverse \cite{prasad1992generalized}, which bring us the best-fit solution in terms of  least squares. The parameters \(p\) serve as weights of particular equations in the system of Eq. \eqref{80} and the solution of Eq. \eqref{81} is equivalent to a weighted least square solution of sysem Eq. \eqref{80}. 

From that, using the Moor-Penrose inverse \cite{prasad1992generalized}, the best-fit parameters change rate is as follows:

\begin{equation}
\label{82}
    \dot{\theta} = P(\theta),
\end{equation}
where
\begin{equation}
\label{83}
    P(\theta) = \left(\grad_\theta y \cdot diag(p) \cdot \left(\grad_\theta y\right)^T\right)^{-1} \cdot \grad_\theta y \cdot diag(p) \cdot f
\end{equation}
is a vector function representing the parameter change rate.

Next, we assume, that the relation \eqref{83} is a gradient of some function  \(\Pi(\theta)\)
\begin{eqnarray}
    \nabla_\theta \Pi(\theta) = P(\theta),
\end{eqnarray}
and we desire to find a maximum of that function. Then, the steady state solution of the Eq. \eqref{82} is satisfied to a maximum of the function \(\Pi(\theta)\).Taking inspiration from Natural Gradient \cite{amari1998natural} decent we can reuse an idea of the decent step in customized distance measure to find the maximum of \(\Pi(\theta)\) and consequently the solution of the original Eq. \eqref{82}.

The gradient accent points in the direction of steepest accent or more formally
\begin{equation}
     d \theta^* = \arg \max \, \Pi(\theta + d \theta) \,\, \text{such as} \,\, ||d \theta ||< \epsilon \,\, \text{for small} \, \epsilon.
\end{equation}

Let the new distance measure be defined as the following function:

\begin{equation}
\label{85}
    ||d\theta||_A= \sqrt{d\theta^T \cdot A^{-1} \cdot d\theta}
\end{equation}

where
\begin{equation}
    A = \grad_\theta y\cdot diag(p)  \cdot\left(\grad_\theta y\right)^T,
\end{equation}
Except the extreme point, with zero gradient terms \( \grad_\theta y\), \(A\) is the positive definite matrix, and so Eq. \eqref{85} is a vector norm \cite{pugh2002real}. 

With the new distance measure, the gradient of the function \(\Pi(\theta)\) is defined as follows: 

\begin{equation}
    \hat{P}(\theta) = A\cdot P(\theta) =  \grad_\theta y \cdot diag(p)\cdot f.
\end{equation}

Then the steady state solution of the original Eq. \eqref{82} is equivalent to the steady state solution of equation 

\begin{equation}
    \dot{\theta} = \hat{P}(\theta)
\end{equation}

describing the gradient accent in the customized distance measure.

Finally, taking the limit w.r.t discretization step we came to the following definition of the  parameter change rate in customized distance measure:

\begin{equation}
    \hat{P}(\theta) = \int p(x) \cdot \grad_\theta y_\theta(x)\cdot f(x,y,\grad_x y) dx.
\end{equation}

We apply the procedure of distance measure replacement to find steady-state solution of value function,  probability density and action policy.

In case of value function optimization, \(x\) is agent state \(s\), \(y\) is value function and

\begin{equation} 
f = \overline{v}\cdot \nabla_s V(s)  + \overline{r} - |\log {\gamma} | \,  V(s) =  \E_{a \sim \pi(a|s)} A(s,a),
\end{equation}

according to  Eq. \eqref{eq:dve_de}. Since \(p\) is arbitrary,  we replace it with arbitrary probability distribution \(\rho\). Then, the later equation may be rewritten as follows:

\begin{equation}
    \hat{P}(\theta) = \int \rho(s) \cdot \grad_\theta V \E_{a \sim \pi(a|s)} A(s,a) ds = \left< \E_{a \sim \pi} \left[\grad_{\theta}{V (s)} \cdot A(s,a) \right]\right>_{\rho}.
\end{equation}

Time-discretized solution of the equation \(\dot{\theta} = \hat{P}(\theta)\) leads to the same scheme as approximate gradient decent (see Eq. \eqref{eq:d_value}) 

\begin{equation}
    {\bf \theta}_{k+1} =\theta_k + \beta \left< \E_{a \sim \pi} \left[\grad_{\theta}{V (s)} \cdot A(s,a) \right]\right>_{\rho}.
\end{equation}

In case of agents distribution, \(x\) is agent state \(s\), \(y\) is probability density and

\begin{equation} 
f =\log \gamma \,\left( \overline{p}(s)  - p_0(s) \right) - \nabla_s   \overline{p}(s) \left< v\right>_{\pi}=G^{\pi}(\overline{p}(s),s),
\end{equation}

according to  Eq. \eqref{ptt2}. Since \(p\) is arbitrary,  we replace it with arbitrary probability distribution \(A \cdot \rho(s)/\overline{p}(s)\), where are is normalizing factor. Then, the later equation may be rewritten as follows:

\begin{equation}
    \hat{P}(\theta) = \int A \cdot \frac{\rho(s)}{\overline{p}(s)} \cdot \grad_\theta \overline{p}(s) \cdot G^{\pi}(\overline{p}(s),s) ds = \left< \E_{a \sim \pi} \left[\grad_{\theta}{\log \overline{p} (s)} \cdot G(\overline{p}(s),s) \right]\right>_{\rho},
\end{equation}

where constant \(A\) is omitted.

Time-discretized solution of the equation \(\dot{\eta} = \hat{P}(\eta)\) leads to the same scheme as approximate gradient decent (see Eq. \eqref{77})

\begin{equation}
    {\bf \eta}_{k+1} =\eta_k + \beta^{P} \left< \grad_\eta \log \overline{p}(s) \cdot G(\overline{p}(s),s) \right>_{\rho}, 
\end{equation}

\end{document}